\documentclass{article} % For LaTeX2e
\usepackage{preprint,times}
\usepackage{geometry}
\usepackage{amsmath,amsfonts,bm}

\def\eqref#1{equation~\ref{#1}}
\def\1{\bm{1}}

\newcommand{\valid}{\mathcal{D_{\mathrm{valid}}}}

\DeclareMathAlphabet{\mathsfit}{\encodingdefault}{\sfdefault}{m}{sl}
\SetMathAlphabet{\mathsfit}{bold}{\encodingdefault}{\sfdefault}{bx}{n}

\usepackage{hyperref}
\usepackage{url}
\usepackage{wrapfig}
\usepackage{graphicx}

\usepackage{subcaption}
\usepackage{algorithm}
\usepackage{algpseudocode}
\usepackage{amsmath}
\usepackage{amssymb}
\usepackage{xcolor}
\usepackage[table]{xcolor}
\usepackage{arydshln}
\usepackage{booktabs}
\usepackage[table]{xcolor}

\title{Inference-Time Projection for Physically Valid Biomolecular Diffusion Models}

\author{
  Qurat-ul-ain\textsuperscript{1}, 
  Yee Whye Teh\textsuperscript{1}, 
  Charlotte M. Deane\textsuperscript{1},
   Matteo Cagiada\textsuperscript{1,2}\\[0.5em] %
  \textsuperscript{1}Department of Statistics,  University of Oxford \\
  \textsuperscript{2}Department of Biology, University of Copenhagen \\[0.5em] %
   \texttt{matteo.cagiada@bio.ku.dk}
}

\begin{document}

\maketitle

\begin{abstract}
    AlphaFold 3-style cofolding models predict biomolecular complexes with high structural accuracy, yet a large fraction of their outputs are physically invalid: chains overlap at interfaces, ligand bond lengths and angles are distorted, rings are non-planar, and stereocentres are inverted. Current approaches either steer the sampler with physics-informed potentials, which multiplies sampling cost and memory overhead making inference impossible on large complexes, or finetune the model, costing time and tying the fix to one architecture. We observe that, unlike structural accuracy, physical validity is fully verifiable at inference time from quantities the sampler already holds.
    % such as van der Waals radii, the molecular graph of a ligand or its distance-geometry bounds. 
    We therefore treat physical validity as a constrained inference problem and introduce two closed-form projection operators applied to the diffusion model's denoised clean-coordinate estimate, $\hat{x}_0$: an inter-chain van der Waals projection that pushes apart the most severely clashing atom pairs, and a ligand distance-geometry projection that restores bond lengths, angles, internal contacts, planarity and chirality. Both operators are local, sparse and displacement-capped, require no network evaluations, gradients or importance sampling, and leave the denoiser and its weights untouched, so they can be dropped into any AF3-style sampler without retraining. Applied to two independently developed models, Boltz-2 and OpenFold-3, across five benchmarks (CASP15, CASP16, the PoseBusters monomer and complex sets, and the Boltz physical-validity test set), our method recovers perfect physical validity while preserving structural-accuracy and ligand-placement metrics. These gains are achieved with negligible runtime and memory overhead, providing a practical, model-agnostic route to physically valid all-atom structure prediction.
\end{abstract}

% Together these
% results suggest that AlphaFold-3-style models already sample close to the
% physically valid manifold, and that a minimal, local, inference-time correction
% is enough to reach it.

%  using bounded local corrections, minimally perturbing the model prediction while enforcing steric feasibility. 
 
\section{Introduction}
Biomolecular structure prediction has moved from single-chain protein modelling toward the prediction of complete molecular assemblies containing proteins, nucleic acids, small molecules, ions, and modified residues. AlphaFold 3 (AF3) \citep{af3} and related cofolding models \citep{boltz2, protenix, chai,of3} have made this possible by framing structure prediction as conditional all-atom generation: given sequences and ligand identities, multiple sequence alignment (MSA), template-derived information, and token- and atom-level features, a learned diffusion model samples the three-dimensional coordinates of the full complex. This shift has substantially broadened the scope of structure prediction, enabling a single model class to encode protein–protein interactions,  protein–nucleic-acid interactions, protein–ligand binding, and a wide range of chemically heterogeneous systems involving ions, modified residues, glycans, and covalent modifications. As a result, these models are rapidly becoming broader engines for structural biology and drug discovery. 

\begin{wrapfigure}{H}{0.48\textwidth}
    \centering
    \vspace{-0.5em}
    \includegraphics[width=0.46\textwidth]{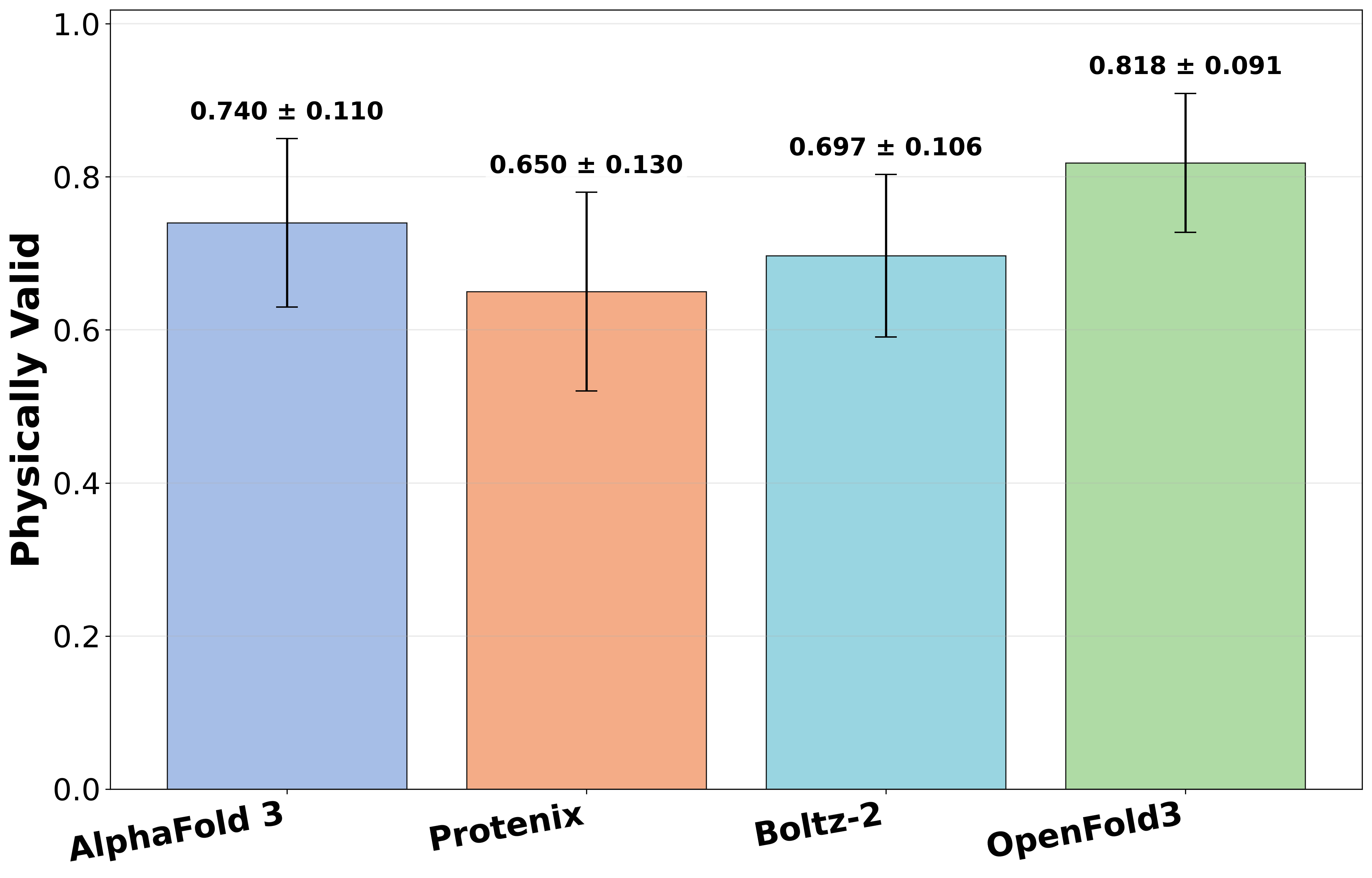}
    \caption{\textbf{Physical validity performance of AlphaFold 3 family models on CASP15.} Oracle (best of 5 samples) per target ($n=66$). Bars show mean across targets; error bars indicate 95\% bootstrap confidence intervals.}
    \label{fig:physical_validity_overview}
    \vspace{-1em}
\end{wrapfigure}

%For a predicted complex to be useful in drug discovery, structural accuracy
%alone is not sufficient; it must also be \textit{physically valid}. 
For a predicted complex to be useful in downstream tasks such as docking, affinity prediction, molecular dynamics, and structure-based optimization, structural accuracy alone is not sufficient; it must also be \textit{physically valid} \citep{physics,Baker2019,posebusters,md,Domagalski2014-gy}.
Following Boltz~\citep{boltz1}, we define
physical validity as the absence of basic all-atom geometry violations: generated
coordinates should satisfy stereochemical constraints, preserve allowable local
ligand geometry, avoid steric clashes, and maintain a physically plausible
organization of the complex. Details of violation classes considered
in this work are given in Appendix \ref{app:physical_violations}. These failures are not rare in AF3-style structure predictors: for 26\% of CASP15 targets, AF3 produces no physically valid structure in any of five samples (Fig.~\ref{fig:physical_validity_overview}). A breakdown of physical validity failures in Boltz-2 \citep{boltz2} and OpenFold3 \citep{of3} is given in Table \ref{tab:validity-failure-modes}. Further evaluation of these models, likewise, reports persistent steric and stereochemical limitations in generated complexes \citep{Ishitani2025-ed,physics}. When these constraints are violated, structures can exhibit strained bond lengths, invalid angles, steric clashes, distorted side-chain packing, chirality inversions, or overlapping chains and interfaces. These defects can corrupt binding-pocket geometry, misplace interaction partners, and make otherwise accurate-looking predictions unreliable for downstream tasks. %such as docking, affinity prediction, molecular dynamics, and structure-based optimization \citep{physics,Baker2019,posebusters,md,Domagalski2014-gy}.

% Earlier protein structure prediction models often operated in torsion-angle, residue-frame, or otherwise constrained representations which imposed substantial geometric structure on the space of allowable conformations \citep{trrosetta,rosettafold,rosettafoldna,af1}. In such parameterizations, much of the local geometry is specified implicitly through internal coordinates or reconstructed from residue-level frames, so the model explores a narrower and more chemically structured space. In contrast, 
Compared to previous structure prediction methods which operated on representations that enforce local chemistry by construction \citep{pointcloud, pointcloud2, backbone, 3dframes}, the current state-of-the-art cofolding models diffuse on raw atom coordinates where physical validity is not imposed as a strict constraint during training. Aside from a bond loss applied to bonded ligands and glycans, AF3 and related models do not optimize physical-validity losses in the diffusion objective. As a result, physically realistic local geometry is encouraged only indirectly through the data distribution, making it possible for globally accurate predictions to remain locally invalid. 

In AF3, physical failures are handled only partially and after sampling: inter-chain clashes are reflected as a high penalty in the confidence score,  down-ranking problematic samples; however, the structure produced itself is not corrected.  Another strategy is post-hoc relaxation, where predicted coordinates are energy-minimized before downstream use; AlphaFold2 \cite{af1}, for example, used post-prediction gradient descent in the Amber force field \cite{amber} to enforce peptide-bond geometry and remove stereochemical violations. However, relaxation is straightforward only for proteins; other molecule types require additional force fields and accurate ligand parameterisation, making its application to all atom biomolecular structure prediction impractical at large scale. 

To our knowledge, only two methods attempt to directly
correct physical invalidity in biomolecular diffusion models. The first is
Boltz-2x ~\citep{boltz1,boltz2}, which improves physical plausibility at inference time
using Feynman--Kac steering~\citep{fksteering}. This approach introduces
physics-informed potentials over seven constraint families, uses importance
sampling over multiple candidates, and applies gradient-based refinement to bias
diffusion toward more valid structures. The second approach is the Gauss--Seidel Projection (GSP) method ~\citep{gsp},
which trains a differentiable projection layer to enforce physical validity as a hard constraint. GSP finetunes Boltz with a differentiable projection module built on the same
seven physical constraint families used by Boltz steering.
% ; accordingly,
% we also develop our method within the Boltz-2 framework to make comparison with both methods as fair as possible.
While GSP offers fast two-step sampling and validity guarantees, it traded accuracy for efficiency and required end-to-end finetuning of the base model.

% We apply a lightweight projection step during inference that maps each denoised coordinate estimate to the nearest configuration satisfying van der Waals non-overlap constraints.  

In this work, we ask: \textit{Can biomolecular diffusion models be made
physically valid without costly pretraining, finetuning, or
compromising speed and structural accuracy?} 
 We summarize our contributions as follows:

\begin{itemize}
\item \textbf{A model-agnostic projection on the clean-coordinate estimate:}
  We introduce two closed-form projection operators applied to the denoised
  estimate $\hat{x}_0$ before each reverse update: an inter-chain van der
  Waals projection that removes steric clashes, and a ligand
  distance-geometry projection that restores bond lengths, angles, internal
  contacts, planarity and chirality. We define physical validity as a verifiable, inference-time constraint, so projection to a valid set can be applied in any diffusion sampler without retraining. We demonstrate this on two
  independently developed AF3-style models, Boltz-2 and OpenFold-3.

\item \textbf{Physical Validity for free:} Our method lifts
  physical validity to $100\%$ on five extensive benchmarks, CASP15, CASP16,
  PoseBusters monomers and complexes, and the Boltz test set, for both base
  models (Table~\ref{tab:main_results}) while preserving
  structural-accuracy and ligand position metrics. There is no trade-off between validity, accuracy and efficiency; every constraint is evaluated in closed form, resulting in negligible wall-clock overhead and no measurable increase in GPU memory.

\end{itemize}
    
\subsection{Related Work}
\paragraph{Constrained and projected diffusion.}
Inference-time guidance steers pretrained diffusion models without retraining
by adding gradients of differentiable losses, rewards, or energies to the
reverse process \citep{lossguided, chung2022improving, energyguided, universalguidance}.
These are soft procedures and are strongly dependent on the guidance weight. If the
weight is too weak, the constraint is not satisfied; if it is too strong,
sample quality drops and samples drift off the data manifold
\citep{trust_sampling, cfg++}. Posterior samplers such as DPS
\citep{dps} make this especially clear, because they perturb $x_t$
using a loss evaluated on a denoised proxy, and an unreliable proxy or an
overly large step pushes the trajectory off-manifold. Later work recasts
guidance as a sequence of local optimization problems. The update is
restricted to the denoiser-induced manifold \citep{chung2022improving}, stopped
once the proxy becomes unreliable \citep{trust_sampling}, or bounded to stay
near the learned trajectory \citep{cfg++}. A second line of work replaces soft guidance with explicit feasibility
corrections. Projected Diffusion Models \citep{constrainedsynthesis} treat reverse
diffusion as constrained optimization and project iterates onto the feasible
set. Predict--Project--Renoise (PPR) \citep{predit_project} projects through the
denoiser, i.e.\ on $\hat{x}_0$, and alternates projection with re-noising.
\citet{proteinconstraint} apply a similar predict--correct--renoise scheme
to protein design, enforcing stereochemical and pocket constraints with an
ADMM-based proximal step. In each case the projection is itself an iterative
optimization inside every reverse step, which adds substantial cost per
sample. Closest to our setting, \citet{gsp} enforce physical validity
in biomolecular cofolding with a Gauss--Seidel projection over all-atom
coordinates. However, this approach requires fine-tuning the diffusion module; applying it only at inference degrades
structural accuracy \cite{gsp}.

Our method keeps the $\hat{x}_0$-space correction of PPR-style samplers, but
replaces the inner solver with a single closed-form, displacement-bounded
correction. This correction is applied once per denoising step during the final,
low-noise portion of sampling; the
bounded corrections accumulate across these steps and drive samples toward
physical validity. A detailed account of contemporary methods in biomolecular diffusion models is
provided in Appendix ~\ref{app:related_works}.

%  

% \section{Background and Notation}

\section{Methods}
% \subsection{Projected Diffusion Algorithm}
\label{app:diffusion_algo}

\newcommand{\hl}[1]{{\color{red}#1}}   % preamble
\begin{algorithm}[h]
\caption{\textsc{ProjectedSampleDiffusion}}
\label{alg:projected_sample_diffusion}
\begin{algorithmic}[1]
\Require $\{f^\ast\}, \{s_i^{\mathrm{inputs}}\}, \{s_i^{\mathrm{trunk}}\}, \{z_{ij}^{\mathrm{trunk}}\}$,
  noise schedule $[c_0,c_1,\dots,c_T]$, \hl{$\mathcal{B}_{\mathrm{proj}}$},
  $\gamma_0 = 0.8$, $\gamma_{\min}=1.0$, noise scale $\lambda = 1.003$, step scale $\eta = 1.5$
\State $\vec{x}_l \sim c_0 \cdot \mathcal{N}(\vec{0}, I_3)$ \hfill $\vec{x}_l \in \mathbb{R}^3$
\For{$c_\tau \in [c_1,\dots,c_T]$}
  \State $\{\vec{x}_l\} \leftarrow \textsc{CentreRandomAugmentation}(\{\vec{x}_l\})$

  \State \label{alg:line:gamma} $\hat{\gamma} \leftarrow \gamma_0 \;\textbf{if}\; c_\tau > \gamma_{\min} \;\textbf{else}\; 0$
\State \label{alg:t}  
$\hat{t} \leftarrow c_{\tau-1}(1+\hat{\gamma});
\textcolor{red}{\quad t_{\tau} \leftarrow 1 - (\tau-1)/T}$
\State \label{alg:line:xi} $\vec{\xi}_l \leftarrow \lambda \sqrt{\hat{t}^2 - c_{\tau-1}^2}\,\mathcal{N}(\vec{0}, I_3)$ \hfill $\vec{\xi}_l \in \mathbb{R}^3$
  \State \label{alg:line:noisy} $\vec{x}^{\mathrm{noisy}}_l \leftarrow \vec{x}_l + \vec{\xi}_l$
  \State \label{alg:line:denoise} $\{\vec{x}^{\mathrm{denoised}}_l\} \leftarrow
    \textsc{DiffusionModule}\big(\{\vec{x}^{\mathrm{noisy}}_l\}, \hat{t}, \{f^\ast\},
    \{s_i^{\mathrm{inputs}}\}, \{s_i^{\mathrm{trunk}}\}, \{z_{ij}^{\mathrm{trunk}}\}\big)$

  \If{\hl{$\textsc{ProjectionActive}({t_{\tau}})$}} \label{alg:line:active}
    \State \label{alg:line:project} \hl{$\{\tilde{\vec{x}}^{\mathrm{denoised}}_l\} \leftarrow
      \Pi_{\mathcal{C}(\mathcal{B}_{\mathrm{proj}})}\big(\{\vec{x}^{\mathrm{denoised}}_l\}\big)$}
  \Else
    \State \label{alg:line:skip} \hl{$\{\tilde{\vec{x}}^{\mathrm{denoised}}_l\} \leftarrow \{\vec{x}^{\mathrm{denoised}}_l\}$}
  \EndIf
  \State \label{alg:xnoisy} $\vec{\delta}_l \leftarrow
    (\vec{x}^{\mathrm{noisy}}_l - \hl{\tilde{\vec{x}}^{\mathrm{denoised}}_l})/\hat{t}$
  \State $dt \leftarrow c_{\tau} - \hat{t}$
  \State \label{alg:line:euler} $\vec{x}_l \leftarrow \vec{x}^{\mathrm{noisy}}_l + \eta\,dt\,\vec{\delta}_l$
\EndFor
\State \Return $\{\vec{x}_l\}$
\end{algorithmic}
\end{algorithm}
Our starting point is an observation about what makes physical validity
different from structural accuracy. Accuracy can only be scored against an
experimental reference that is unavailable at inference time, but validity is
defined entirely by quantities an AF3 sampler already has such as van der Waals radii,
the molecular graph of a ligand and its RDKit \cite{rdkit} distance-geometry bounds, ring
planarity and assigned stereochemistry. Every violation is therefore
exactly verifiable on any coordinate estimate at any step. We treat physical validity as a constrained
inference problem and exploit this verifiability with a lightweight projection
on the diffusion model's denoised clean-coordinate estimate,
$\hat{x}_0$. 

% Diffusion models generate samples by iteratively transforming a noisy state
% \(\mathbf{x}_t\) into a cleaner state. In unconstrained generation, this reverse
% process is learned only from the data distribution. However, in protein structure prediction,
% just producing high-likelihood samples is not sufficient: generated structures must
% also satisfy physical constraints such as local distance
% geometry, stereochemistry, and chain non-overlap. We therefore view structure
% sampling as a constrained generation problem, where the goal is to remain close
% to the learned structural distribution while reducing physically invalid
% configurations.
Algorithm~\ref{alg:projected_sample_diffusion} is a modified version of the
AlphaFold~3 sample-diffusion procedure (Algorithm~18 in the Supplementary
Information of \citet{af3}); our modifications are shown in red. Here,
$\mathcal{B}_{\mathrm{proj}}$ denotes the batch-level structural metadata used
to construct the projection constraints: atom masks, atom-to-token mappings,
chain and entity indices, bonded connectivity, elements (used to assign van der
Waals radii), and, for ligands, RDKit-derived distance-geometry bounds for
bonded, angle-implied and non-neighbouring atom pairs, together with masks
indicating each constraint subtype.

At each reverse step, the diffusion module maps the noised coordinates
(line~\ref{alg:line:noisy}) to a denoised estimate of the clean all-atom
structure (line~\ref{alg:line:denoise}). The sampler converts this estimate
into a reverse drift (line~\ref{alg:xnoisy}) and takes an Euler step from
$\vec{x}^{\mathrm{noisy}}_l$ toward it (line~\ref{alg:line:euler}).  We make a small, displacement bound correction to
$\vec{x}^{\mathrm{denoised}}_l$ (i.e. $\hat{x}_0$), which changes the direction of the subsequent drift (line~\ref{alg:xnoisy}). When $\textsc{ProjectionActive}(t_{\tau}) = \mathbf{1}[t_{\tau} \leq
\texttt{projection\_start\_t}]$ holds, where $t_{\tau}$ is the normalised sampling time
of Appendix~\ref{sec:algorithms}, the raw estimate is
replaced by its projection $\Pi_{\mathcal{C}(\mathcal{B}_{\mathrm{proj}})}$
(lines 9--10); otherwise it passes through unchanged (line~\ref{alg:line:skip}). 

% Because the drift is computed
% from the projected estimate, the correction steers the sampling trajectory
% itself rather than being applied after sampling.

$\Pi_{\mathcal{C}(\mathcal{B}_{\mathrm{proj}})}$ is a bounded local correction
operator composed of two projections applied in sequence to the same denoised
estimate (Appendix~\ref{sec:algorithms}). The van der Waals projection
(Algorithm~\ref{alg:vdw_projection}) targets inter-chain steric clashes: it
evaluates candidate atom pairs across chains against their van der Waals lower
bounds, selects the $K$ most deeply penetrating pairs, and applies a symmetric
push-apart to each. The ligand distance-geometry (LDG) projection
(Algorithm~\ref{alg:ldg_projection}) corrects the internal geometry of each ligand
toward the constraints implied by its molecular graph: bonded and angle pairs
are held within RDKit-derived distance intervals, non-bonded intra-ligand pairs
are kept above a clash lower bound, aromatic rings and double-bond groups are
flattened toward a fitted plane, and sp$^3$ stereocentres whose signed volume
disagrees with the assigned handedness receive a Newton step that re-inverts
them. Both operators accumulate the contributions of all selected violated
constraints involving atom $l$ into a single displacement $\vec{D}_l$ and cap
only the sum, $\tilde{\vec{x}}^{\mathrm{denoised}}_l =
\vec{x}^{\mathrm{denoised}}_l + \vec{D}_l$ where $\|\vec{D}_l\|_2 \le
\delta_{\max}$, so an atom involved in many violations is bounded on its net
motion rather than on each contribution. The cap is applied once per call of
the van der Waals projection and once per pass of the ligand projection.
Neither operator evaluates or differentiates through the network or modifies
its weights; each correction is computed in closed form from the current
denoised coordinates and $\mathcal{B}_{\mathrm{proj}}$. In practice, the corrections are small and shrink over the tail: the
RMS and maximum per-atom displacement decrease over the final 40 steps while
the mean van der Waals penetration depth falls toward zero, so projection acts
as a sparse local repair rather than a global rearrangement of the structure
(Appendix~\ref{app:diffusion_projection_diagnostics}).

\paragraph{When and where to project:}
\label{when}
The AlphaFold~3 sampler re-injects noise at every step with $c_\tau >
\gamma_{\min} = 1$\,\AA{} (``churn'' \cite{edm}, lines~\ref{alg:line:gamma}--\ref{alg:line:noisy}).
% :the noise level is raised to $\hat{t} = 1.8\,c_{\tau-1}$ and Gaussian noise
% with per-coordinate standard deviation $\approx 1.5\,c_{\tau-1} \ge 1.5$\,\AA{}
% is added. 
Once $c_\tau \le \gamma_{\min}$, no fresh noise is injected, and the
remaining steps are deterministic. Two observations
(Appendix~\ref{app:sampler_dynamics}) shape our design.

First, the structure is essentially fixed by the time churn switches off (step 154 out of 200). The
denoised estimate $\hat{x}_0$ comes within $1$\,\AA{} of the final structure
just before the churn boundary, then drops to $\sim$$0.1$\,\AA{}, and its
LDDT levels off (Fig.~\ref{fig:traj_summary}). In the deterministic tail the
denoiser therefore moves atoms by less than $0.1$\,\AA{}, well below the
projection cap $\delta_{\max} = 0.45$\,\AA{}, so projected corrections are
not overwritten and accumulate. During churn, each step injects noise an order
of magnitude larger than the correction, which erases it. This is visible
directly in Fig.~\ref{fig:vdw_violations}: projecting only during churn undoes the projection due to high noise and violation count remains similar to no projection baseline. This is also observed in (Table~\ref{tab:sen_position}) where projection only during churn leaves physical validity at $0.02$ whereas projection during the tail reduces violations steadily and reaches $1.00$.

Second, steric constraints are meaningful on a clean structure. Until
the tail, $x_t$ sits at a distance of $\approx\sigma$ from the final structure
and has near-zero LDDT, whereas $\hat{x}_0$ is a coherent structure from the
first step (Figs.~\ref{fig:traj_rmsd}, \ref{fig:traj_lddt}). We therefore
project $\hat{x}_0$ and activate projection only in
the tail ($\texttt{projection\_start\_t} = 0.2$, i.e.\ from step $\approx 160$). Once churn is off, $x_t$ and $\hat{x}_0$ differ by less than $\sigma \le
1$\,\AA{}, and projecting either gives the same validity and accuracy
(Table~\ref{tab:sen_position}).

\section{Results}

\subsection{How projection interacts with the learned prior}
\label{sec:penetration-trajectory}

The projection is local, sparse, and displacement-limited: it moves only the atoms involved in the most
severe violations; pushing clashing pairs apart, pulling over-stretched bonds
together, flattening distorted rings and re-inverting flipped stereocentres, while
capping the maximum per-atom movement. Whether such a scheme
can reach physical validity depends on how the denoiser responds to each
correction: if the network simply reverts it at the next step, nothing
accumulates; if some of it is retained, the effect compounds. To measure this
directly we instrument the Boltz-2 sampler and track inter-chain penetration
along the trajectory.

\begin{figure}[t]
  \centering
  \includegraphics[width=\linewidth]{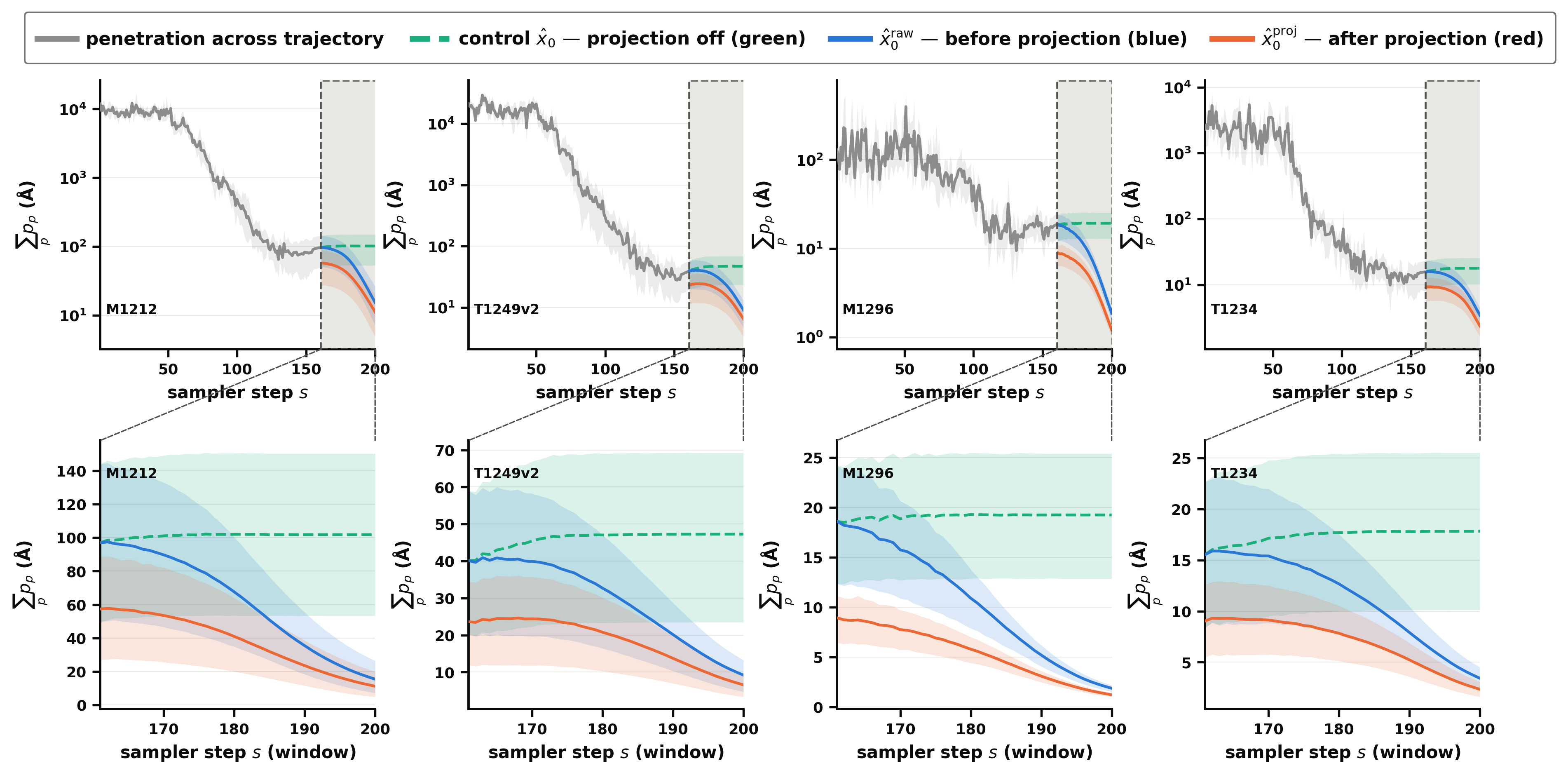}
  % \caption{%
  %   \textbf{Inter-chain penetration along the Boltz-2 reverse trajectory} for
  %   the four CASP16 targets with the highest projection activity; five samples each.
  %   \textbf{Left:} total penetration $P$ (Eq.~\ref{eq:penetration}) of the
  %   sampler state over all 200 steps (log scale); the shaded region marks the
  %   40-step projection window.
  %   \textbf{Right:} the projection window on a linear axis, showing
  %   $\hat{\mathbf{x}}_0$ from the seed-matched control run without projection
  %   (green, dashed), $\hat{\mathbf{x}}_0^{\mathrm{raw}}$ before projection
  %   (blue), and $\hat{\mathbf{x}}_0^{\mathrm{proj}}$ after projection (red).
  %   Lines are means over 5 trajectories; bands are 95\% bootstrap
  %   intervals.
  %   The blue--red gap at a step is what one projection removes; the green--blue
  %   gap is the correction the model has internalized. The control stays flat
  %   while the model's own unprojected estimate converges toward the projected
  %   one.%
  % }
    \caption{%
    \textbf{Inter-chain penetration along the Boltz-2 reverse trajectory} for the
    four most projection-active CASP16 targets, five samples each.
    \textbf{Top:} total penetration over all 200
    steps (log scale); shading marks the 40-step projection window.
    \textbf{Bottom:} that window on a linear axis, showing
    $\hat{\mathbf{x}}_0$ for the seed-matched control (green, dashed),
    $\hat{\mathbf{x}}_0^{\mathrm{raw}}$ before projection (blue) and
    $\hat{\mathbf{x}}_0^{\mathrm{proj}}$ after (red); means over 5 trajectories,
    bands 95\% bootstrap intervals. The blue--red gap is what one projection
    removes; the green--blue gap is what the model has internalized.%
  }
  \label{fig:penetration-trajectory}
\end{figure}

\paragraph{Setup:}
% For a coordinate set $\mathbf{x}$ we define the total inter-chain penetration
% \begin{equation}
%   P(\mathbf{x}) \;=\; \sum_{p \in \mathcal{P}} p_p(\mathbf{x}),
%   \quad
%   p_p(\mathbf{x}) = \big[\, \ell_p - \| \mathbf{x}_i - \mathbf{x}_j \| \,\big]_{+},
%   \quad
%   \ell_p = (1-b)\,(r_i + r_j),
%   \quad p = (i,j)
%   \label{eq:penetration}
% \end{equation}
% where $\mathcal{P}$ is the set of inter-chain heavy-atom pairs considered by
% Algorithm~\ref{alg:vdw_projection}, $r_i, r_j$ are the van der Waals radii,
% $\ell_p$ is the buffered lower bound enforced by the projection with buffer
% $b = 0.15$, and $[\cdot]_{+} = \max(\cdot,0)$. $P$ is the summed penetration
% depth in \AA{} over all pairs violating their bound and vanishes if and only
% if no inter-chain pair does; unlike a binary validity label, it resolves the
% severity of clashes, not only their presence.
For a coordinate set $\mathbf{x}$, $P(\mathbf{x}) = \sum_{(i,j)\in\mathcal{P}}
\big[\ell_{ij} - \|\mathbf{x}_i-\mathbf{x}_j\|_2\big]_+$ is the total inter-chain
penetration over the candidate pairs of Algorithm~\ref{alg:vdw_projection}, with
$\ell_{ij}=(1-\beta)(r_i^{\mathrm{vdw}}+r_j^{\mathrm{vdw}})$ and $\beta=0.15$.
We selected four CASP16 targets with the highest projection activity and
ran five samples each, giving 20 trajectories. At every step $s$ of the
projection window $s \in [161, 200]$ we log $P$ on the denoiser's clean
estimate at three points: before projection,
$\hat{\mathbf{x}}_0^{\mathrm{raw}}(s)$; immediately after projection,
$\hat{\mathbf{x}}_0^{\mathrm{proj}}(s) = \Pi\big(\hat{\mathbf{x}}_0^{\mathrm{raw}}(s)\big)$;
and in the model's next prediction one step later,
$\hat{\mathbf{x}}_0^{\mathrm{raw}}(s{+}1)$. Comparing the first two isolates
what the projection removes at step $s$,
% \begin{equation}
%   \Delta_s \;=\; P\big(\hat{\mathbf{x}}_0^{\mathrm{raw}}(s)\big) - P\big(\hat{\mathbf{x}}_0^{\mathrm{proj}}(s)\big) ;
% \end{equation}
comparing the last two measures how much of the removed overlap the model
re-introduces in its next prediction, which we call the \emph{rebound}: with $\Delta_s = P(\hat{\mathbf{x}}_0^{\mathrm{raw}}(s))
- P(\hat{\mathbf{x}}_0^{\mathrm{proj}}(s))$,
$\rho_s = \big[P(\hat{\mathbf{x}}_0^{\mathrm{raw}}(s+1)) -
P(\hat{\mathbf{x}}_0^{\mathrm{proj}}(s))\big] / \Delta_s$ and
retention $= 1-\rho_s$. As a reference we run the identical sampler with the projection disabled
(same seeds, same hyper-parameters) and record $\hat{\mathbf{x}}_0$ over the
same window (\emph{control}); any divergence between the control and
$\hat{\mathbf{x}}_0^{\mathrm{raw}}$ is therefore attributable to the projection
alone. Figure~\ref{fig:penetration-trajectory} shows the full trajectories and
the projection window; confidence intervals below are 95\% bootstrap intervals
over the 20 trajectories.

\paragraph{Each correction is small and absorbed by the learned prior.}
Averaged over the window, the model retains around ${\sim}8\%$ of each
correction at each step and re-expresses the rest from its own learned geometry
(rebound $\rho = 0.919$, 95\% CI $[0.915, 0.925]$). This is by design: the
per-step correction is deliberately small and displacement-capped, intended to
nudge the model repeatedly rather than override it in one shot. The projection
is iterative and its effect is cumulative: the fraction retained at each step,
though small, is reinforced at every subsequent step, and across the 40-step
window this steadily carries the structure to validity.

\paragraph{The trajectory internalizes the corrections.} Measuring
penetration on $\hat{\mathbf{x}}_0^{\mathrm{raw}}$, i.e.\ on the structure the
model proposes on its own before we touch it, the penetration of the model's
unprojected predictions falls by $80.3\%$ across the window
(95\% CI $[77.5\%, 83.0\%]$; per-target trajectories in
Fig.~\ref{fig:penetration-trajectory}, bottom). Over the same steps the seed-matched control shows no reduction at all: its penetration
rises by $12.0\%$ (95\% CI $[9.1\%, 14.9\%]$), so this
convergence is produced by our method. The model's own geometry becomes nearly clash-free i.e. it has
internalized the corrections rather than being held apart by force at the final
step. This is also what distinguishes in-loop projection from a post-hoc
correction applied once to the final coordinates
(Appendix ~\ref{app:posthoc}).

% \item \textbf{We characterize the dynamics of projected sampling and show that the denoiser absorbs, rather than undoes, the correction.} Evaluating the denoising trajectory at each step, we show that the projection does not hold structures in place by force: while the denoiser reverses roughly $91\%$ of each step's edit, the fraction it retains grows through the window and compounds, so that
%   the model's own unprojected predictions become nearly clash-free before
%   sampling ends. This absorption behaviour is what allows a small, capped
% per-step correction to achieve full validity: the work is spread across the
% many steps rather than concentrated in a single large, trajectory-disrupting
% move.

% Decomposing the trajectory into projection work
% and denoiser response, we find that the denoiser's own next estimate retains a
% large and stable fraction of each correction 

\subsection{Experimental Setup}

We evaluate each method along three complementary axes:  physical
validity, which measures whether the predicted all-atom coordinates satisfy
basic stereochemical and steric constraints; structural accuracy,
which measures agreement with the experimental reference structure; and computational efficiency, measured
as wall-clock time and peak GPU memory per sample. We evaluated five benchmark sets: CASP15, CASP16, the
PoseBusters monomer and complex sets, and Test, the physical-validity dataset
released with Boltz-1 \cite{boltz1}. We apply our method to two
independently developed AF3-style base models, Boltz-2 \citep{boltz2} and
OpenFold-3 \cite{of3}, and compare each projected model against its
unmodified base. We additionally run Boltz-2x i.e. Boltz-2 with gradient-based FK steering \citep{fksteering} for physical validity, for comparison with an existing inference-time method.
Boltz-2x exhausts an 80\,GB A100 on nine targets: H1236, R1283v2 and R1285
(CASP16); 7b2c, 7cnq, 7suc, 8aql and 8g0v (PoseBusters complexes); and 8sro
(Test). All nine complete under the base model and reach full physical
validity under our projection. A direct comparison to GSP \cite{gsp} was not possible due to a lack of publicly released model checkpoints. Further details of the experimental set up are given in Appendix~\ref{app:experimental_setup}.

\subsection{Accuracy and Validity}
\begin{table}[t]
\centering
\caption{Structural quality and physical validity, oracle over 5 samples, with non-parametric bootstrap standard errors (10,000 resamples over targets). DockQ$_{>0.23}$ is the fraction of scored interfaces with DockQ above 0.23, and LRMSD$_{<2}$ the fraction of ligands placed within 2\,\AA. DockQ requires an interface and LDDT-PLI/LRMSD a scored ligand. PoseBusters monomers have no interfaces, hence no DockQ.}
\label{tab:main_results}
\footnotesize
\setlength{\tabcolsep}{5pt}
\begin{tabular}{@{}l r@{$\pm$}l r@{$\pm$}l r@{$\pm$}l r@{$\pm$}l r@{$\pm$}l@{}}
\toprule
Method & \multicolumn{2}{c}{LDDT$\uparrow$} & \multicolumn{2}{c}{DockQ$_{>0.23}\uparrow$} & \multicolumn{2}{c}{LDDT-PLI$\uparrow$} & \multicolumn{2}{c}{LRMSD$_{<2}\uparrow$} & \multicolumn{2}{c}{Phys.\ Valid$\uparrow$} \\
\midrule\midrule
\multicolumn{11}{c}{\textbf{CASP15}} \\
\midrule
\textit{n} & \multicolumn{2}{c}{66} & \multicolumn{2}{c}{14} & \multicolumn{2}{c}{12} & \multicolumn{2}{c}{12} & \multicolumn{2}{c}{66} \\
\midrule
Boltz-2 & 0.457 & 0.038 & 0.786 & 0.109 & 0.564 & 0.085 & 0.453 & 0.126 & 0.697 & 0.056 \\
\textbf{Ours (Boltz)} & 0.456 & 0.038 & 0.786 & 0.110 & 0.571 & 0.083 & 0.453 & 0.127 & 1.000 & 0.000 \\
Boltz-2x & 0.457 & 0.038 & 0.714 & 0.119 & 0.524 & 0.092 & 0.425 & 0.131 & 1.000 & 0.000 \\
OpenFold-3 & 0.440 & 0.039 & 0.821 & 0.096 & 0.510 & 0.075 & 0.408 & 0.128 & 0.818 & 0.048 \\
\textbf{Ours (OpenFold)} & 0.440 & 0.039 & 0.821 & 0.096 & 0.514 & 0.076 & 0.408 & 0.129 & 1.000 & 0.000 \\
\midrule\midrule
\multicolumn{11}{c}{\textbf{CASP16}} \\
\midrule
\textit{n} & \multicolumn{2}{c}{52} & \multicolumn{2}{c}{18} & \multicolumn{2}{c}{9} & \multicolumn{2}{c}{9} & \multicolumn{2}{c}{52} \\
\midrule
Boltz-2 & 0.453 & 0.044 & 0.500 & 0.103 & 0.371 & 0.062 & 0.363 & 0.076 & 0.538 & 0.069 \\
\textbf{Ours (Boltz)} & 0.452 & 0.044 & 0.500 & 0.103 & 0.372 & 0.061 & 0.363 & 0.076 & 1.000 & 0.000 \\
Boltz-2x & 0.451 & 0.044 & 0.519 & 0.097 & 0.372 & 0.063 & 0.322 & 0.081 & 1.000 & 0.000 \\
OpenFold-3 & 0.449 & 0.044 & 0.583 & 0.093 & 0.332 & 0.053 & 0.237 & 0.104 & 0.692 & 0.064 \\
\textbf{Ours (OpenFold)} & 0.449 & 0.044 & 0.583 & 0.093 & 0.332 & 0.053 & 0.237 & 0.104 & 1.000 & 0.000 \\
\midrule\midrule
\multicolumn{11}{c}{\textbf{PoseBusters Monomer}} \\
\midrule
\textit{n} & \multicolumn{2}{c}{134} & \multicolumn{2}{c}{0} & \multicolumn{2}{c}{134} & \multicolumn{2}{c}{134} & \multicolumn{2}{c}{134} \\
\midrule
Boltz-2 & 0.954 & 0.003 & \multicolumn{2}{c}{--} & 0.741 & 0.024 & 0.687 & 0.040 & 0.948 & 0.019 \\
\textbf{Ours (Boltz)} & 0.954 & 0.003 & \multicolumn{2}{c}{--} & 0.741 & 0.024 & 0.687 & 0.040 & 1.000 & 0.000 \\
Boltz-2x & 0.954 & 0.003 & \multicolumn{2}{c}{--} & 0.727 & 0.024 & 0.649 & 0.041 & 1.000 & 0.000 \\
OpenFold-3 & 0.951 & 0.003 & \multicolumn{2}{c}{--} & 0.830 & 0.019 & 0.817 & 0.033 & 0.858 & 0.030 \\
\textbf{Ours (OpenFold)} & 0.951 & 0.003 & \multicolumn{2}{c}{--} & 0.831 & 0.019 & 0.810 & 0.034 & 1.000 & 0.000 \\
\midrule\midrule
\multicolumn{11}{c}{\textbf{PoseBusters Complex}} \\
\midrule

\textit{n} & \multicolumn{2}{c}{164} & \multicolumn{2}{c}{161} & \multicolumn{2}{c}{164} & \multicolumn{2}{c}{164} & \multicolumn{2}{c}{164} \\
\midrule
Boltz-2 & 0.908 & 0.008 & 0.628 & 0.036 & 0.689 & 0.022 & 0.621 & 0.037 & 0.500 & 0.039 \\
\textbf{Ours (Boltz)} & 0.908 & 0.008 & 0.628 & 0.036 & 0.689 & 0.022 & 0.621 & 0.037 & 1.000 & 0.000 \\
Boltz-2x & 0.907 & 0.008 & 0.630 & 0.036 & 0.687 & 0.022 & 0.603 & 0.037 & 1.000 & 0.000 \\
OpenFold-3 & 0.909 & 0.008 & 0.602 & 0.037 & 0.777 & 0.020 & 0.731 & 0.034 & 0.585 & 0.038 \\
\textbf{Ours (OpenFold)} & 0.909 & 0.007 & 0.602 & 0.038 & 0.778 & 0.020 & 0.731 & 0.034 & 1.000 & 0.000 \\
\midrule\midrule
\multicolumn{11}{c}{\textbf{Test}} \\
\midrule
\textit{n} & \multicolumn{2}{c}{541} & \multicolumn{2}{c}{344} & \multicolumn{2}{c}{251} & \multicolumn{2}{c}{251} & \multicolumn{2}{c}{541} \\
\midrule
Boltz-2 & 0.831 & 0.006 & 0.714 & 0.022 & 0.646 & 0.022 & 0.619 & 0.028 & 0.608 & 0.021 \\
\textbf{Ours (Boltz)} & 0.831 & 0.006 & 0.714 & 0.022 & 0.645 & 0.022 & 0.619 & 0.028 & 1.000 & 0.000 \\
Boltz-2x & 0.834 & 0.006 & 0.707 & 0.022 & 0.649 & 0.022 & 0.623 & 0.027 & 1.000 & 0.000 \\
OpenFold-3 & 0.816 & 0.007 & 0.686 & 0.023 & 0.599 & 0.023 & 0.594 & 0.027 & 0.678 & 0.020 \\
\textbf{Ours (OpenFold)} & 0.816 & 0.007 & 0.686 & 0.022 & 0.599 & 0.023 & 0.594 & 0.027 & 1.000 & 0.000 \\
\bottomrule
\end{tabular}
\end{table}

\textbf{Our method achieves perfect physical validity without sacrificing structural
accuracy.} 
Table~\ref{tab:main_results} shows that our method achieves 100\% pass rates on physical validity on every dataset with structural accuracy unchanged. The two projection operators (vdW and LDG) are applied to two independently trained samplers, so the correction is model-agnostic. Accuracy differences between Boltz-2 and Boltz-2x are within noise at these sample sizes; Appendix~\ref{sec:qualitatitve} shows a failure mode steering can produce: Boltz-2x
collapses the five sampled poses onto a single, physically clean, but incorrect
binding mode.
% The picture is the same for OpenFold-3, which our operator brings physical validity to $100\%$ with every accuracy value almost exactly retained. 

Because the projection moves atoms, we verified in
Appendix~\ref{app:structural_effects} that validity is not bought at the
expense of structure quality elsewhere. Three independent analyses agree. (i) Applying
projection after sampling instead of during inference either leaves most
violations in place (one pass removes $14\%$ of targeted violations) or, when
iterated to feasibility, significantly lowers LDDT; post-hoc Phenix \citep{phenix}
minimization reaches only $0.881$ validity while costing $0.022$ LDDT, with
the largest losses on the most accurate structures
(\S\ref{app:posthoc}). (ii) Covalent geometry is essentially unchanged: bond and angle
RMSD from ideal stay within $0.0002$\,\AA{} and $0.015^\circ$ of Boltz-2 and angle RMSD remains closer to ideal than the deposited experimental structures
(\S\ref{app:covalent}). 
% The one metric that moves is the worst-case bond deviation (median 0.114{\AA}  versus 0.070{\AA} for Boltz-2), whose upper tail is nonetheless lower with projection (p95 0.568{\AA} versus 0.625{\AA}) (iii) Backbone and side-chain conformational outlier
% rates are unchanged ($0.17\%$ Ramachandran outliers for both) and well below
% experimental rates (\S\ref{app:conformations}). 
(iii) Packing is preserved:
after an identical short relaxation, projected and base structures reach the
same Lennard-Jones energy on every target, and the number of contacts in the
solvent-accessible alpha complex is unchanged at the median ($\le 0.01\%$), with the largest per-target change below 2\%
(\S\ref{app:packing}). The projection therefore removes physical violations
rather than redistributing them into other geometric defects. Unlike fine-tuned projection \citep{gsp}, it preserves accuracy without
changing any weights; unlike post-hoc refinement, it corrects the trajectory
while the denoiser can still adapt to the correction; and unlike
inference-time steering (Boltz-2x), it adds no measurable runtime or memory
(\S\ref{sec:runtime_memory}).

\subsection{Efficiency}
\label{sec:runtime_memory}
\begin{figure}[t]
  \centering
  \captionsetup[subfigure]{skip=1pt}
  \begin{subfigure}[b]{\linewidth}
    \centering
    \includegraphics[width=\linewidth,trim=3mm 3mm 3mm 3mm,clip]{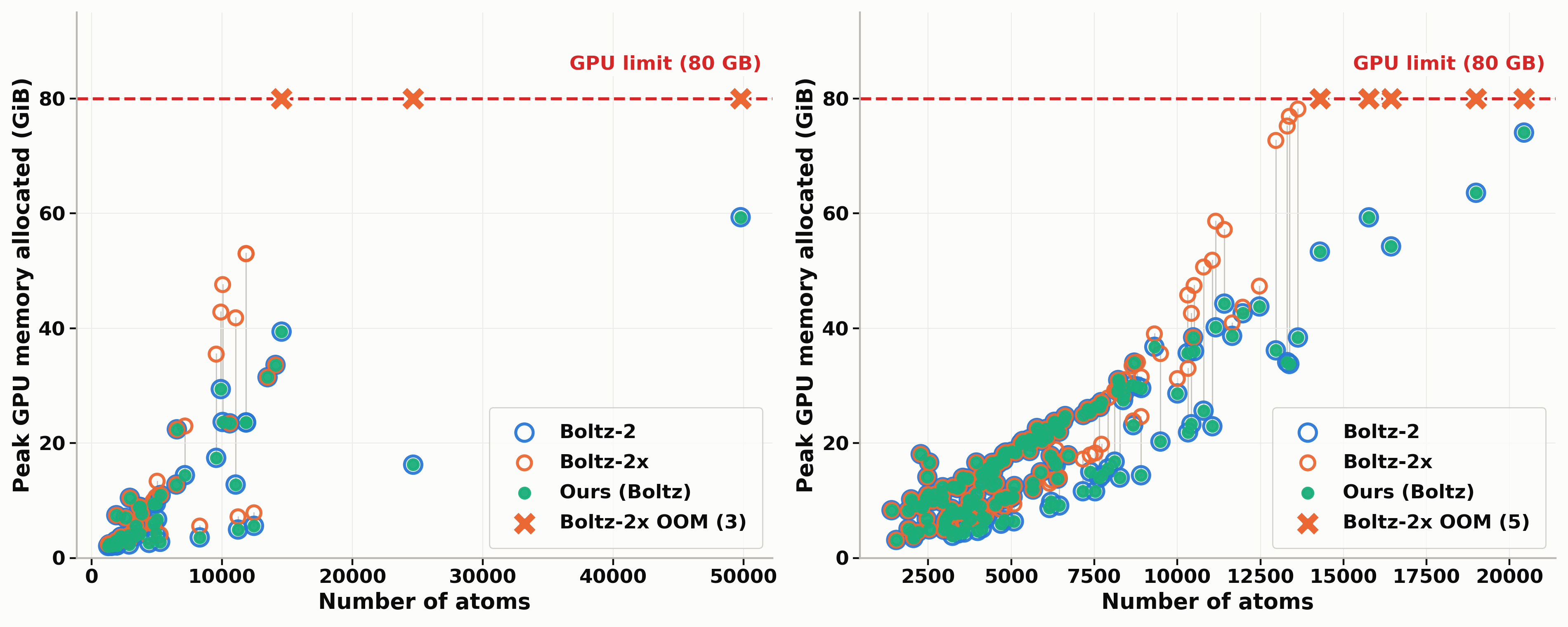}
    \caption{Peak GPU memory versus structure size.}
    \label{fig:mem_size}
  \end{subfigure}%
  \\[0.15em]
  \begin{subfigure}[b]{\linewidth}
    \centering
    \includegraphics[width=\linewidth,trim=3mm 3mm 3mm 3mm,clip]{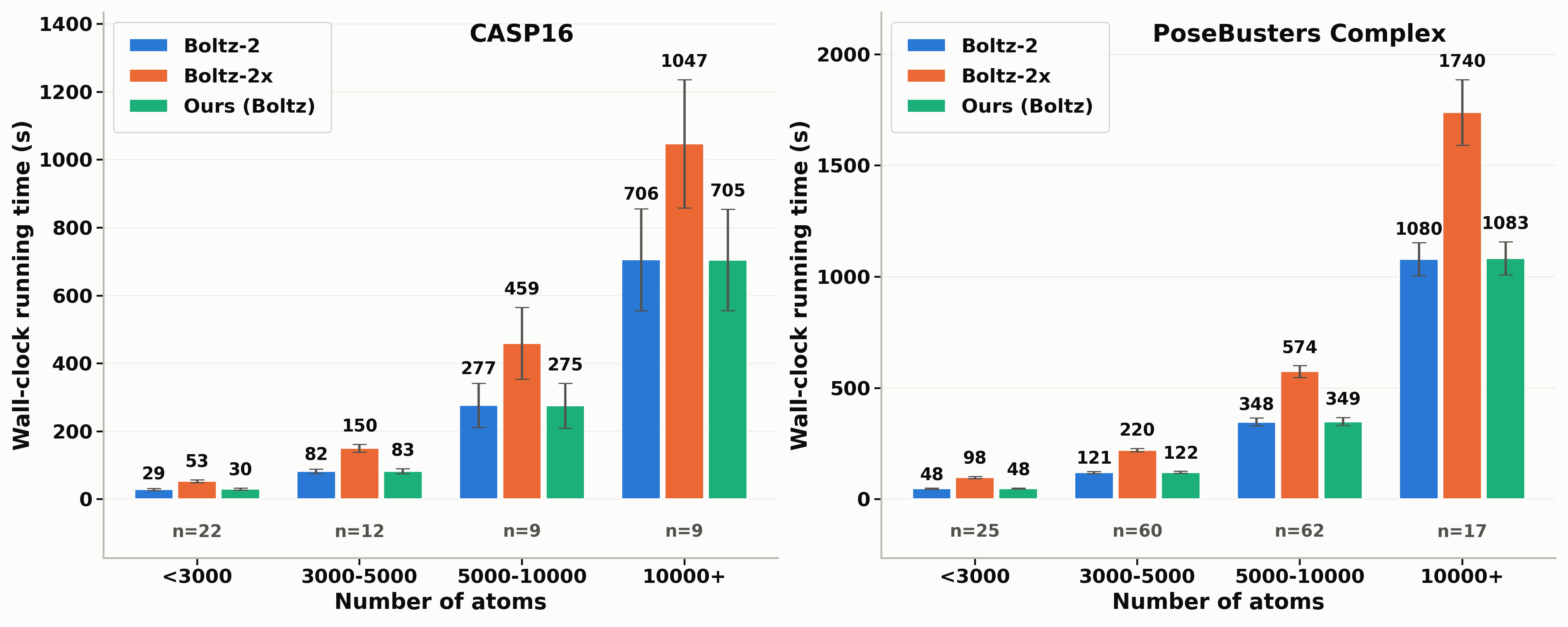}
    \caption{Inference time stratified by atom-count bin.}
    \label{fig:runtime}
  \end{subfigure}
  \caption{\textbf{Cost of inference-time physical correction on CASP16 and
  PoseBusters Complex.} \textbf{(a)}~Peak PyTorch CUDA memory per target; thin grey
  lines join the three methods on the same target. \textbf{(b)}~Mean wall-clock time
  per target by atom count; error bars are the standard error of the mean, $n$ the
  number of targets per bin.}
  \label{fig:runtime-time}
\end{figure}

\textbf{Our method achieves perfect physical validity at no measurable time or memory overhead.} Our central motivation is that physical validity should be
improved without substantially increasing sampling cost.
Figure~\ref{fig:mem_size} shows that our method adds no memory overhead: the
base model and our method coincide over the full size range on both
benchmarks as projection adds no network evaluations and no gradients. Steering is expensive at inference time. Boltz-2x exceeds the 80 GB A100 budget on nine targets, including three CASP16 and five PoseBusters Complex targets (Figure \ref{fig:mem_size}), and Figure \ref{fig:runtime} shows the same picture in wall-clock time. Across all size bins, runtime with projection stays within
$3.5\%$ of the base model, and differences of both signs indicate that any
remaining gap is below run-to-run variation. Boltz-2x costs $1.5$--$2.1\times$
the base model throughout, and its absolute overhead grows steeply with size:
on complexes above $10{,}000$ atoms it adds $341$\,s on CASP16 and $660$\,s on
PoseBusters Complex. This is a
major limitation as AF3-style models are often used in high-throughput
settings to screen many ligands, sample multiple seeds, or generate alternative
structural hypotheses. Our method therefore provides a scalable
and practical route to physically valid all-atom predictions while largely
preserving the speed and memory profile of the base model. A detailed account of the time and memory comparison of both base models on all datasets is given in Table \ref{tab:time_mem_combined}.

\subsection{Ablations}
\begin{table}[t]
\centering
\footnotesize
\caption{Where and when the projection is applied, on 134 CASP15, CASP16 and PoseBusters Monomer and Complex targets on which Boltz-2 produced at least one physically invalid sample. Higher is
better for all metrics. 
% \emph{Position}: $\hat{x}_0$ projects the denoised estimate inside the sampler
% (final configuration), and post hoc applies the same
% operator after sampling, either once or iterated to feasibility (at most 40
% passes). \emph{$x_t$ window}: the $t_{\mathrm{start}}$ sweep repeated with the
% projection applied to $x_t$. \emph{Churn phase only}: the vdW projection runs
% only on the steps that still inject fresh noise (steps 0--154, $\sigma >
% 1$\,\AA) and never on the churn-free tail; the final configuration is the
% reverse, projecting only in the churn-free tail. 
Dashed, shaded row is the
final configuration; red cells are worse than it under a Holm-corrected paired
test.}
\label{tab:sen_position}
\setlength{\tabcolsep}{4pt}
\begin{tabular}{@{}l c c c c c c@{}}
\toprule
Setting & Value & LDDT & DockQ$_{>0.23}$ & LDDT-PLI & LRMSD$_{<2}$ & Phys.\ valid \\
\midrule
\multicolumn{7}{@{}l}{\emph{Projection position}} \\
\hdashline
\rowcolor{green!12}Position & $\hat{x}_0$ & $0.81 \pm 0.02$ & $0.55 \pm 0.04$ & $0.54 \pm 0.06$ & $0.53 \pm 0.08$ & $1.00 \pm 0.00$ \\
\hdashline
Position & $x_t$ & $0.81 \pm 0.02$ & $0.55 \pm 0.04$ & $0.53 \pm 0.06$ & $0.53 \pm 0.08$ & $1.00 \pm 0.00$ \\
Position & post hoc, once & $0.81 \pm 0.02$ & $0.55 \pm 0.04$ & $0.53 \pm 0.06$ & $0.52 \pm 0.08$ & \cellcolor{red!15}$0.22 \pm 0.04$ \\
Position & post hoc, iterated & \cellcolor{red!15}$0.79 \pm 0.02$ & $0.55 \pm 0.04$ & $0.53 \pm 0.06$ & $0.52 \pm 0.08$ & $0.90 \pm 0.01$ \\
% \midrule
% \multicolumn{7}{@{}l}{\emph{$x_t$ window ($t_{\mathrm{start}}$ applied to the noisy iterate)}} \\
% $x_t$: $t_{\mathrm{start}}$ & 0.01 & $0.81 \pm 0.02$ & $0.55 \pm 0.04$ & $0.53 \pm 0.06$ & $0.52 \pm 0.08$ & \cellcolor{red!15}$0.22 \pm 0.04$ \\
% $x_t$: $t_{\mathrm{start}}$ & 0.05 & \cellcolor{red!15}$0.80 \pm 0.02$ & $0.55 \pm 0.04$ & $0.54 \pm 0.06$ & $0.53 \pm 0.08$ & $1.00 \pm 0.00$ \\
% $x_t$: $t_{\mathrm{start}}$ & 0.20 & $0.81 \pm 0.02$ & $0.55 \pm 0.04$ & $0.54 \pm 0.06$ & $0.53 \pm 0.08$ & $1.00 \pm 0.00$ \\
% $x_t$: $t_{\mathrm{start}}$ & 0.40 & $0.81 \pm 0.02$ & $0.55 \pm 0.04$ & $0.54 \pm 0.06$ & $0.53 \pm 0.07$ & $1.00 \pm 0.00$ \\
% $x_t$: $t_{\mathrm{start}}$ & 0.80 & $0.81 \pm 0.02$ & $0.55 \pm 0.04$ & $0.54 \pm 0.06$ & $0.53 \pm 0.08$ & $1.00 \pm 0.00$ \\
% $x_t$: $t_{\mathrm{start}}$ & 1.00 & $0.81 \pm 0.02$ & $0.55 \pm 0.04$ & $0.54 \pm 0.06$ & $0.53 \pm 0.07$ & $1.00 \pm 0.00$ \\
\midrule
\multicolumn{7}{@{}l}{\emph{Churn phase only (no projection in the churn-free tail)}} \\
Churn only & $\hat{x}_0$ & $0.81 \pm 0.02$ & $0.54 \pm 0.04$ & $0.53 \pm 0.06$ & $0.53 \pm 0.08$ & \cellcolor{red!15}$0.02 \pm 0.01$ \\
Churn only & $x_t$ & $0.81 \pm 0.02$ & $0.54 \pm 0.04$ & $0.53 \pm 0.06$ & $0.53 \pm 0.08$ & \cellcolor{red!15}$0.04 \pm 0.02$ \\
\bottomrule
\end{tabular}
\end{table}

Tables ~\ref{tab:sen_position} validates the design choices introduced in \S\ref{when}.
Projecting only while churn is active is useless: validity reaches just $0.02$
($\hat{x}_0$) and $0.04$ ($x_t$), because each churn step injects noise an order of
magnitude larger than the correction it overwrites, so nothing accumulates. Once churn
ends, $x_t$ and $\hat{x}_0$ lie within $\sigma \leq 1$\,\AA{} of each other and of the
final structure, and the table confirms that in this regime the two are
interchangeable; both reach $1.00$ validity with identical accuracy. What matters is
therefore the window, not which iterate is projected. Moving the same operator outside
the sampler fails in two distinct ways. A single post-hoc pass removes only $14\%$ of
the targeted violations and reaches $0.22$ validity, because each pass is
displacement-capped and one is not enough to attain feasibility in one go. Applying projection after sampling multiple times does attain it ($0.90$), but there is no longer a denoiser to absorb the
corrections: they are imposed on the final coordinates instead of being integrated into
the trajectory, and LDDT falls from $0.81$ to $0.79$. In-loop projection is what buys
both, feasibility and an intact structure, and this is the same absorption
mechanism measured directly in \S\ref{sec:penetration-trajectory}. 
Table~\ref{tab:sensitivity} ablates each hyperparameter of the two operators.

\section{Discussion and Limitations}

Our results suggest that physical-validity failures in AF3-style cofolding
models are not, for the most part, failures of global structural reasoning.
They are small residual violations that remain after the learned denoiser has
already placed the molecular components in approximately correct positions. On
the benchmarks we study, a bounded inference-time projection toward a
physically valid constraint set is enough to remove these failures while
leaving the structural prior of the base model intact. Physical correction
therefore does not necessarily require retraining, expensive relaxation, or
diffusion-time steering with multiple learned potentials. 
% The denoiser can
% instead be treated as a strong proposal mechanism for globally plausible
% structures, with a lightweight constrained operator enforcing local
% feasibility on the clean-coordinate estimate before it is propagated through
% the sampler. 
More broadly, this points to a practical division of labour for
biomolecular diffusion models: learned generative models supply global
structural hypotheses, while small constrained inference operators enforce the
local physical constraints that are hard to guarantee from data fitting alone.
Such predictions can then support not only global structural hypotheses but
also atomistic reasoning about binding, specificity, designability and
molecular mechanism, which requires structures that are locally correct as
well as globally plausible.

% Because every constraint is evaluated in closed form, the correction is also
% cheap. It requires no additional network evaluations, no gradients through the
% denoiser, and no importance sampling over candidate structures, with cost linear in
% the number of constraints and no learned parameters.
% In this way, the sampler receives just
% enough correction to remove violations without disrupting the learned
% denoising trajectory.

\textbf{Limitations:} Our operator is a bounded partial projection, not a hard satisfaction onto
the feasible set. Each step moves an atom by at most $\delta_{\max}$, and only
the $K$ most deeply penetrating pairs are corrected per step, so validity
is reached empirically rather than guaranteed. Moreover, the validity criteria used to evaluate all methods in this study (following \cite{boltz1}) should be
read as evidence that the targeted failure modes are removed, not as a general
certificate of physical realism. We address this with independent
checks: MolProbity covalent geometry, OpenMM van der
Waals energies, and alpha-complex contact analysis
(\S\ref{app:structural_effects}). Additionally, validity is measured against the constraint families that the projection
enforces, inter-chain contacts, ligand bond lengths and angles, internal
clashes, planarity and chirality, so a criterion outside them, such as
torsional strain, protein backbone geometry or hydrogen placement, is neither corrected nor assessed here. 

% Physical
% validity is also not correctness: a structure can be clash-free and still
% have the wrong binding mode or interface, and a local repair operator cannot
% fix a globally wrong prediction.

% At the same time, adding constraints must be done carefully: overly aggressive correction could move samples away from the learned structural manifold and degrade ligand placement or interface accuracy. Future work should therefore study adaptive projection schedules and uncertainty-aware criteria for deciding when and how strongly to project. 

% e (Saad, 2003), which sweeps over all constraints for a few iterations, enforcing each locally
% by updating only the affected atoms. It leverages locality and sparsity of the constraints and thereby
% yields faster, more stable convergence than gradient descent

\bibliography{references}
\bibliographystyle{preprint}

\clearpage
\appendix
\section*{Appendix}

\addcontentsline{toc}{section}{Appendix}

\section{Characterizing Physical Violations}
\label{app:physical_violations}
\begin{figure}[h]
    \centering

    \begin{subfigure}[t]{0.48\linewidth}
        \centering
        \includegraphics[width=\linewidth]{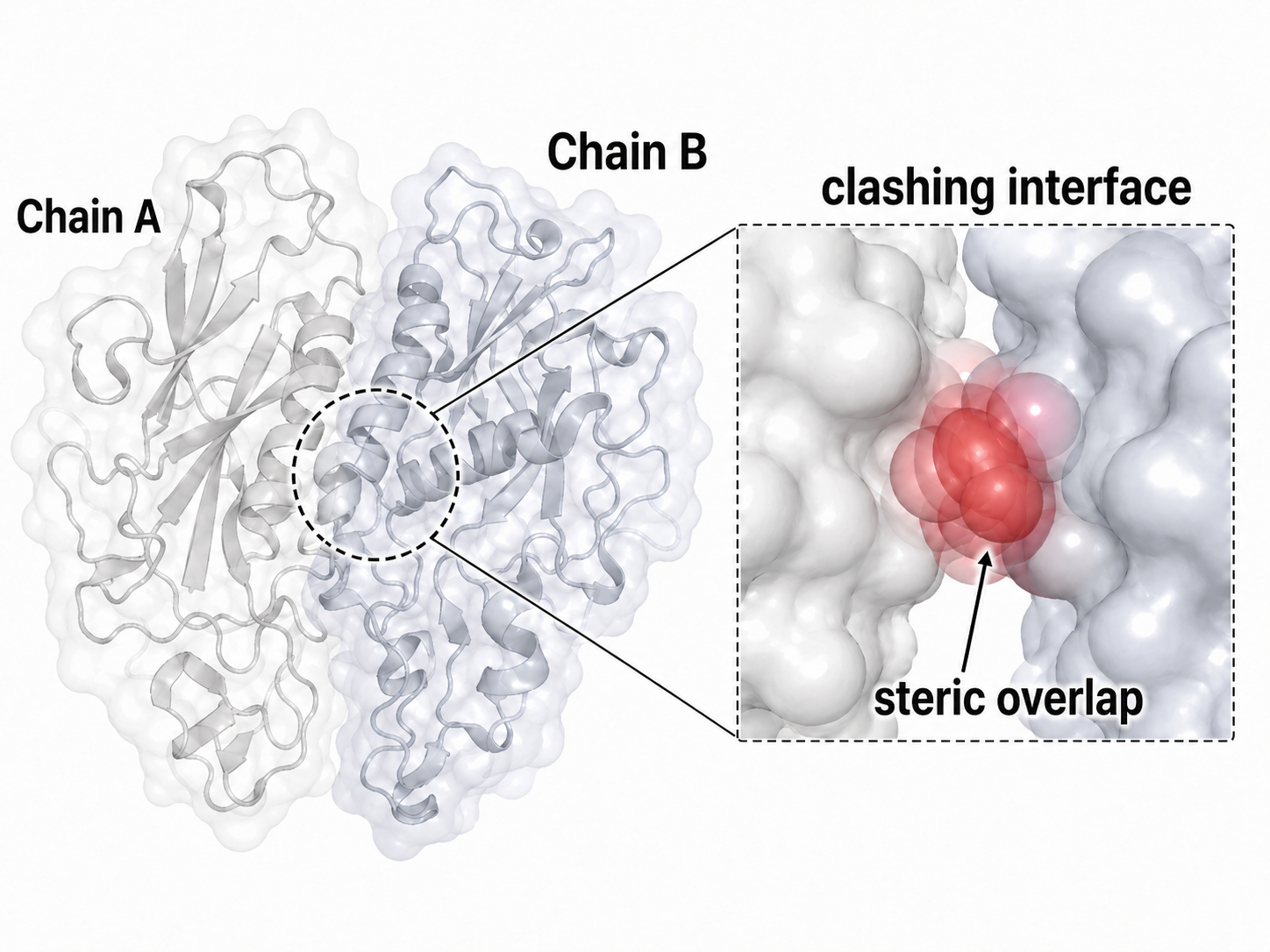}
        \caption{Inter-chain steric clash.}
        \label{fig:violations_chain_clash}
    \end{subfigure}
    \hfill
    \begin{subfigure}[t]{0.48\linewidth}
        \centering
        \includegraphics[width=\linewidth]{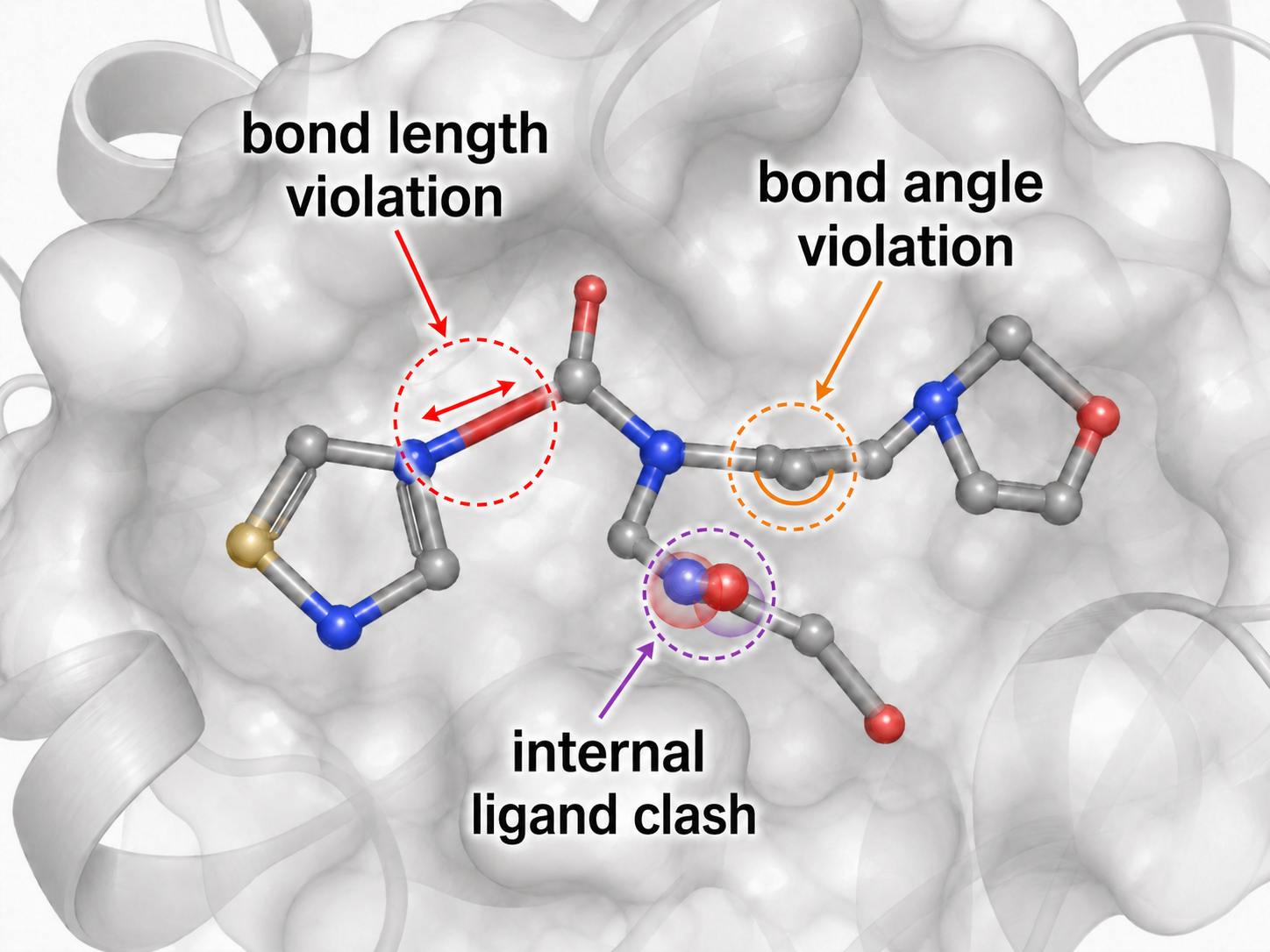}
        \caption{Ligand distance-geometry violations.}
        \label{fig:violations_bonds}
    \end{subfigure}

    \begin{subfigure}[t]{0.48\linewidth}
        \centering
        \includegraphics[width=\linewidth]{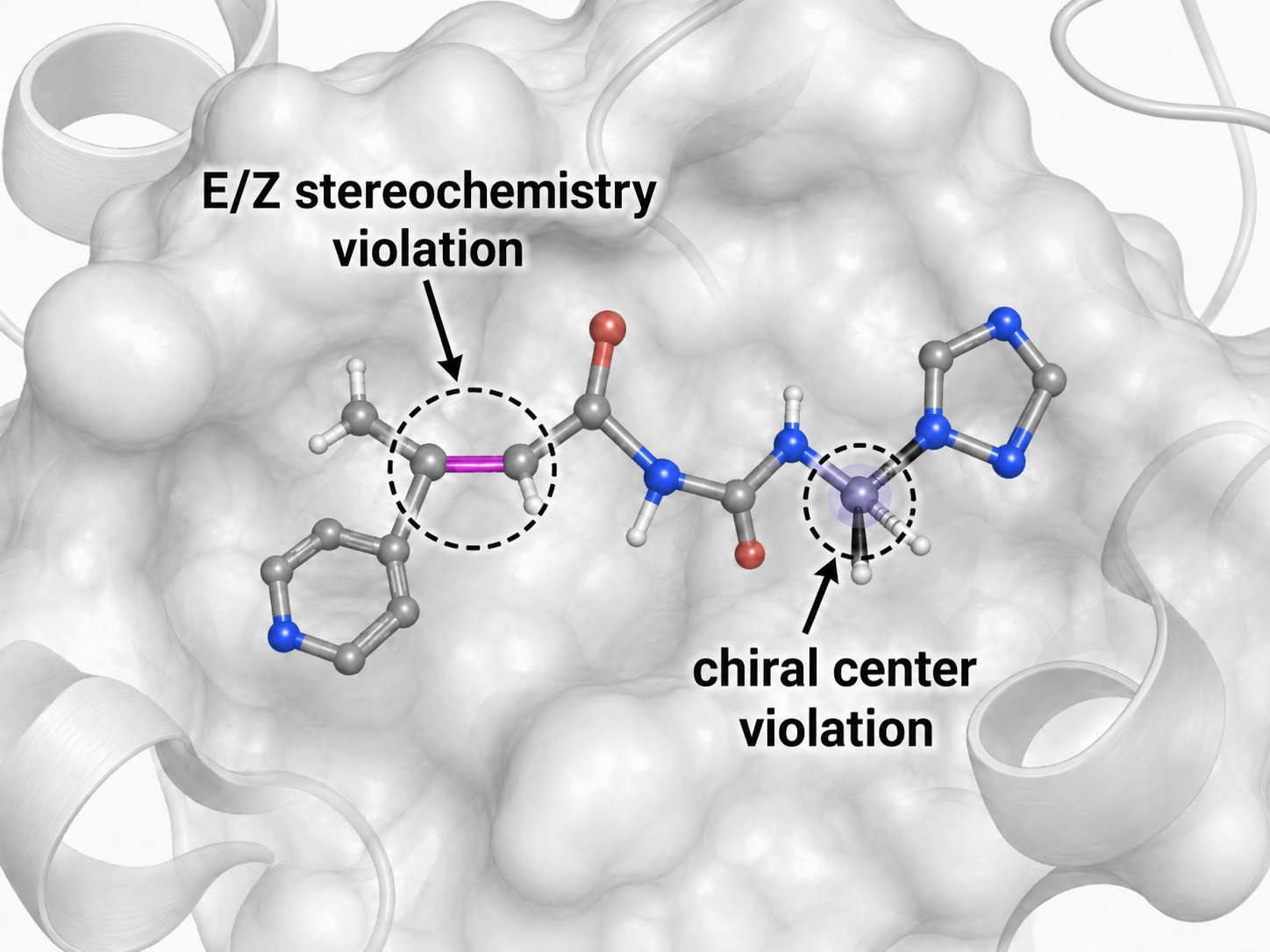}
        \caption{Ligand stereochemistry violations.}
        \label{fig:violations_stereo}
    \end{subfigure}
    \hfill
    \begin{subfigure}[t]{0.48\linewidth}
        \centering
        \includegraphics[width=\linewidth]{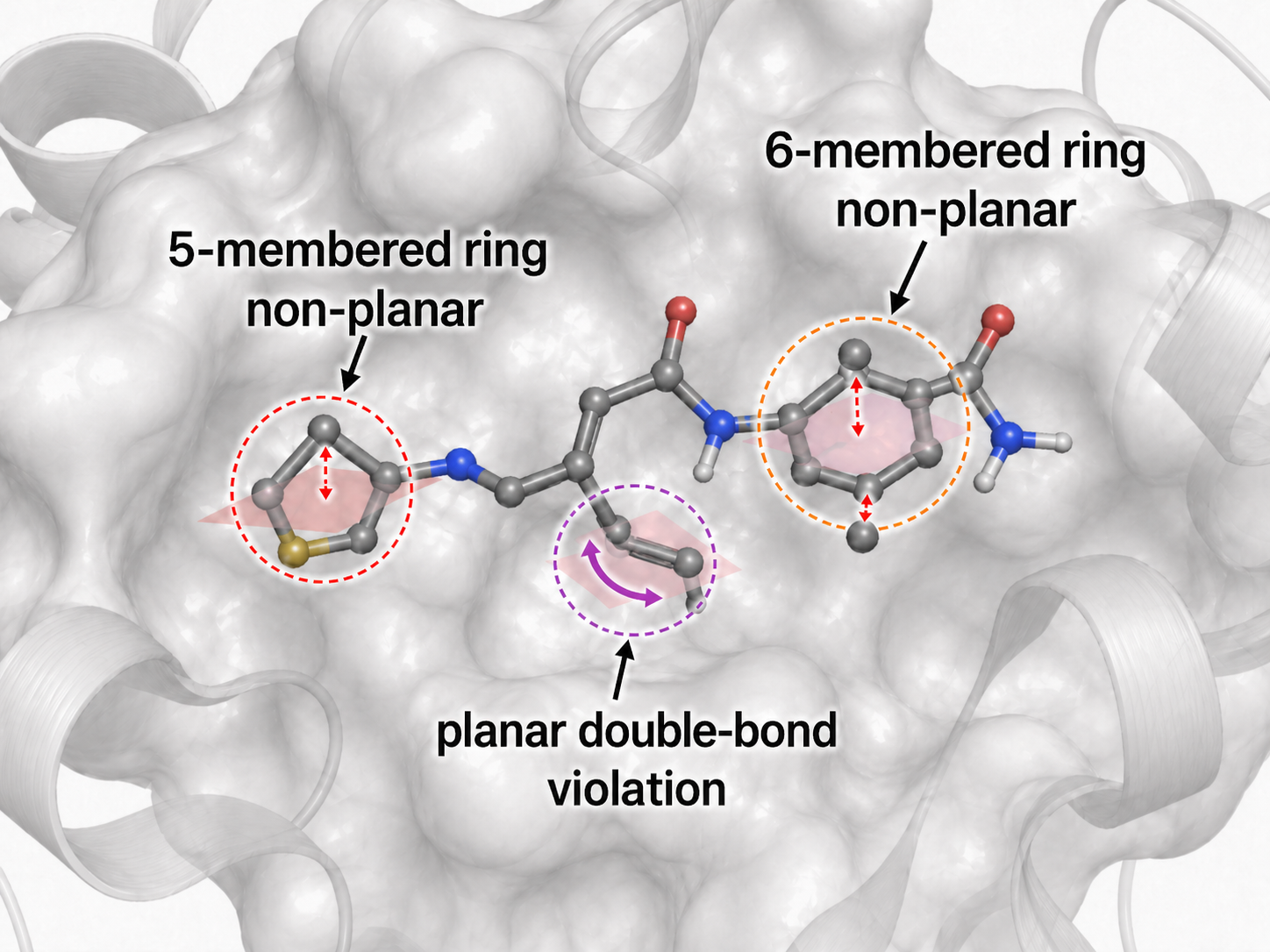}
        \caption{Ligand planarity violations.}
        \label{fig:violations_planar}
    \end{subfigure}

    \caption{
    Examples of physical validity violations in predicted biomolecular complexes.
    \textbf{(a)} Inter-chain steric clash, where two chains overlap at the interface; the zoomed inset highlights atoms occupying incompatible van der Waals volumes.
    \textbf{(b)} Ligand distance-geometry violations, showing invalid bond length, distorted bond angle, and internal ligand clash.
    \textbf{(c)} Ligand stereochemistry violations, including incorrect E/Z double-bond geometry and inverted chirality.
    \textbf{(d)} Planarity violations in aromatic rings and planar double-bond groups.
    }
    \label{fig:physical_violations}
\end{figure}
Figure~\ref{fig:physical_violations} illustrates the main classes of violations considered in this work, set by \cite{boltz1}. These errors can occur even when the overall protein fold or ligand placement appears plausible, and they can compromise downstream use in docking, affinity prediction, molecular dynamics, or structure-based design.

\textbf{Figure~\ref{fig:physical_violations} a: Inter-chain Clashes}: Here, two protein chains are placed so close at the interface that their molecular surfaces overlap. The zoomed inset highlights the local steric overlap, corresponding to atoms from different chains occupying incompatible van der Waals (vdW) volumes. vdW interactions describe the short-range physical constraints between atoms that are not covalently bonded. Each atom can be approximated as occupying a characteristic vdW radius. If two non-bonded atoms are closer than the sum of their vdW radii, their electron clouds overlap, producing a strong repulsive interaction. In our setting, vdW violations, therefore, correspond to steric overlaps: atoms from the ligand, protein, or different chains occupying incompatible volumes in 3D space.

 \textbf{Figure~\ref{fig:physical_violations} b: Ligand Distance-geometry Violations}: A bond length violation occurs when two covalently bonded atoms are placed at an unrealistic separation, either too close or too far apart relative to chemically allowed bounds. A bond angle violation occurs when three connected atoms form an implausible local angle. The same panel also shows an internal ligand clash, where two non-bonded atoms within the ligand occupy overlapping van der Waals volumes.

\textbf{Figure~\ref{fig:physical_violations} c: Stereochemical Violations}: An E/Z stereochemistry violation occurs when the substituents around a double bond are arranged with the wrong cis/trans configuration. A chiral center violation occurs when the tetrahedral arrangement around a stereocentre is inverted, producing the wrong handedness. These errors are particularly important because stereochemistry determines the three-dimensional placement of functional groups in the binding pocket.

\textbf{Figure~\ref{fig:physical_violations} d: Planarity Violations}: Aromatic five- and six-membered rings are expected to remain approximately planar, while double-bond-containing groups such as alkenes and amides should resist twisting out of plane. Deviations from these expected planes indicate strained or chemically implausible ligand conformations.

Table \ref{tab:validity-failure-modes} shows the physical validity failure modes of Boltz-2 and Openfold-3.

\begin{table*}[t]
\centering
\caption{Physical validity failure modes of the base models, Boltz-2 and OpenFold-3. For each target the oracle sample is selected (a physically valid one if any of the five exists, otherwise the one with fewest violations); columns give the total number of violations summed over those structures, and chain clashes the total number of clashing chain pairs. Phys. valid is the percentage of targets for which any of the five samples is valid (oracle). $\cdot$~=~0 (no violations).}
\label{tab:validity-failure-modes}
\footnotesize
\setlength{\tabcolsep}{3pt}
\begin{tabular}{@{}l l cccccccc c@{}}
\toprule
Dataset & Method & \shortstack{Bond\\length} & \shortstack{Bond\\angle} & \shortstack{Internal\\clash} & \shortstack{Chiral-\\ity} & \shortstack{Stereo\\bond} & \shortstack{Planar\\ring} & \shortstack{Planar\\dbl.\ bond} & \shortstack{Chain\\clashes} & \shortstack{Phys.\\valid (\%)} \\
\midrule
\shortstack[l]{CASP15\\($n=66$)} & Boltz-2 & 2 & $\cdot$ & 6 & $\cdot$ & $\cdot$ & $\cdot$ & $\cdot$ & 40 & 69.70 \\
 & OpenFold-3 & $\cdot$ & $\cdot$ & 6 & $\cdot$ & $\cdot$ & $\cdot$ & $\cdot$ & 18 & 81.82 \\
\addlinespace
\shortstack[l]{CASP16\\($n=52$)} & Boltz-2 & $\cdot$ & $\cdot$ & $\cdot$ & $\cdot$ & $\cdot$ & $\cdot$ & $\cdot$ & 47 & 53.85 \\
 & OpenFold-3 & $\cdot$ & $\cdot$ & $\cdot$ & $\cdot$ & $\cdot$ & $\cdot$ & $\cdot$ & 26 & 69.23 \\
\addlinespace
\shortstack[l]{PoseBusters\\Monomer\\($n=134$)} & Boltz-2 & $\cdot$ & $\cdot$ & $\cdot$ & $\cdot$ & $\cdot$ & $\cdot$ & $\cdot$ & 7 & 94.78 \\
 & OpenFold-3 & $\cdot$ & $\cdot$ & $\cdot$ & 20 & $\cdot$ & $\cdot$ & $\cdot$ & 1 & 85.82 \\
\addlinespace
\shortstack[l]{PoseBusters\\Complex\\($n=164$)} & Boltz-2 & $\cdot$ & $\cdot$ & 4 & $\cdot$ & $\cdot$ & $\cdot$ & $\cdot$ & 139 & 50.00 \\
 & OpenFold-3 & $\cdot$ & $\cdot$ & 4 & 56 & $\cdot$ & $\cdot$ & $\cdot$ & 65 & 58.54 \\
\addlinespace
\shortstack[l]{Test\\($n=541$)} & Boltz-2 & 31 & 45 & 10 & $\cdot$ & $\cdot$ & $\cdot$ & $\cdot$ & 576 & 60.81 \\
 & OpenFold-3 & $\cdot$ & $\cdot$ & 2 & $\cdot$ & $\cdot$ & $\cdot$ & $\cdot$ & 494 & 67.84 \\
\bottomrule
\end{tabular}
\end{table*}

\section{Related Work}
\label{app:related_works}
\subsection{Biomolecular Diffusion Models}

Diffusion models for protein modeling and design have explored a range of structural representations, including 3D point clouds \citep{pointcloud,pointcloud2}, backbone internal coordinates \citep{backbone}, and residue-level 3D frames \citep{3dframes}.  Earlier protein structure prediction models often operated in torsion-angle, residue-frame, or otherwise constrained representations which imposed substantial geometric structure on the space of allowable conformations \citep{trrosetta,rosettafold,rosettafoldna,af1}. In such parameterizations, much of the local geometry is specified implicitly through internal coordinates or reconstructed from residue-level frames, so the model explores a narrower and more chemically structured space.  In contrast, current state-of-the-art cofolding models have expanded from backbone-level generation to all-atom modeling and design \citep{protpardelle,rfdiffallatom}. This removes many of these built-in geometric constraints: atoms are generated directly in Cartesian space rather than through
a representation that enforces bonds, angles, steric exclusion, or
stereochemistry by construction. This increased representational flexibility is
essential for modeling heterogeneous complexes with arbitrary ligands and
modified residues, but it also makes physical validity a more explicit inference
problem. 

The current wave of cofolding models is led by AF3, which introduced a diffusion-based all-atom architecture for joint prediction of complexes containing proteins, nucleic acids, ligands, ions, and modified residues, and reported strong gains over specialized baselines across protein--ligand, protein--nucleic acid, and antibody--antigen benchmarks \citep{af3}. AF3 performs prediction by first constructing a trunk representation of the input complex from sequence, MSA, templates, and molecular features, and then using a diffusion module to iteratively denoise noisy all-atom coordinates into a final 3D structure. Open-source models quickly followed this paradigm \citep{boltz1,boltz2,chai,of3,protenix}. Boltz-1 reproduced AF3-style biomolecular interaction modeling in an open framework and introduced physics-inspired steering at inference time \citep{boltz1}. Boltz-2 extended this family with improved structure prediction and an additional binding-affinity prediction module \citep{boltz2}. OpenFold-3 aims to be a faithful reproduction of the AF3 architecture and training recipe \citep{of3}, while Protenix explored a few-step diffusion variant that reduces inference cost while maintaining strong predictive accuracy \citep{protenix}. Together, these models define the current state of the art in all-atom biomolecular cofolding.

\section{Methods}
\label{sec:algorithms}
% ===========================================================================
% Ligand distance-geometry (LDG) projection -- final subsection.
% Order: header + intro -> four algorithms -> constraint construction ->
%        explanatory paragraphs -> sampler integration -> parameters.
% Requires: amsmath, amssymb, algorithm, algpseudocode.
% ===========================================================================

\subsection{Van der Waals Projection}
\label{sec:vdw_projection}      % <- move here, directly after \subsection
\begin{algorithm}[t]
\caption{\textsc{VDWProjection}}
\label{alg:vdw_projection}
\begin{algorithmic}[1]
\Require Denoised coordinates $\hat{X} \in \mathbb{R}^{B \times N_{\mathrm{atom}} \times 3}$;
  batch features $\mathcal{B}_{\mathrm{proj}}$;
  normalised sampling time $t_{\tau} \in (0,1]$
\Require Buffer $\beta$, damping $\alpha$, pair budget $K$,
  displacement cap $\delta_{\max}$, numerical floor $\varepsilon$
\Ensure Projected coordinates $X'$
\If{$t_{\tau} > \texttt{projection\_start\_t}$}
  \State \Return $\hat{X}$ \Comment{projection inactive outside the window}
\EndIf
\State $(\mathcal{P}, \ell) \leftarrow \textsc{BuildVdwArgs}(\mathcal{B}_{\mathrm{proj}})$
  \Comment{$\ell_{ij} = (1-\beta)\,(r^{\mathrm{vdw}}_i + r^{\mathrm{vdw}}_j)$}
\State $X' \leftarrow \mathrm{clone}(\hat{X})$
\For{$b = 1,\dots,B$}
  \State $C \leftarrow X'_b$
  \State $d_{ij} \leftarrow \max(\|C_i - C_j\|_2,\ \varepsilon)$
    \Comment{for all $(i,j) \in \mathcal{P}$}
  \State $p_{ij} \leftarrow \max(\ell_{ij} - d_{ij},\ 0)$ \Comment{penetration depth}
  \State $\mathcal{V} \leftarrow \{(i,j) \in \mathcal{P} : p_{ij} > 0\}$
  \If{$\mathcal{V} = \emptyset$}
    \State \textbf{continue} \Comment{no violation in this batch element}
  \EndIf
  \State $\mathcal{T} \leftarrow \textsc{TopK}_{p}\big(\mathcal{V},\ \min(K, |\mathcal{V}|)\big)$
    \Comment{$K$ most severe violations}
  \State $D \leftarrow 0$ \Comment{per-atom displacement accumulator}
  \ForAll{$(i,j) \in \mathcal{T}$}
    \State $u_{ij} \leftarrow \tfrac{1}{2}\,\alpha\, p_{ij}\, (C_i - C_j)/d_{ij}$
      \Comment{symmetric push-apart}
    \State $D_i \leftarrow D_i + u_{ij}$;\quad $D_j \leftarrow D_j - u_{ij}$
  \EndFor
  \For{$i = 1,\dots,N_{\mathrm{atom}}$}
    \State $D_i \leftarrow D_i \cdot
      \min\!\big(1,\ \delta_{\max} / \max(\|D_i\|_2,\ \varepsilon)\big)$
      \Comment{per-atom displacement cap}
  \EndFor
  \State $X'_b \leftarrow C + D$
\EndFor
\State \Return $X'$
\end{algorithmic}
\end{algorithm}
 
Algorithm~\ref{alg:vdw_projection} is a van der Waals projection layer between the
denoising prediction and the drift computation. Each atom carries an
effective excluded-volume radius, and two non-bonded atoms are not expected to approach closer
than the sum of their radii; distances below this threshold indicate overlapping excluded volumes
and strong steric repulsion, which we treat as clash violations. 

Let $\tau = 1,\dots,T$ index the denoising steps ($T = 200$ in AF3) and let
$t_{\tau} = 1 - (\tau-1)/T$ denote the normalised sampling time, so that $t_1 = 1$ at
the first reverse step and $t_T = 1/T$ at the last. Given denoised all-atom coordinates
$\hat{X} \in \mathbb{R}^{B \times N_{\mathrm{atom}} \times 3}$ and batch-level
structural metadata $\mathcal{B}_{\mathrm{proj}}$ containing the allowed VDW
lower-bound distances, the projection is active over the late part of the reverse
trajectory, whenever $t_{\tau} \leq$ \texttt{projection\_start\_t}; outside that
window $\hat{X}$ is returned unchanged. With $T = 200$ and
$\texttt{projection\_start\_t} = 0.2$ this is the final $40$ steps. 
 
We construct a set of candidate atom pairs
$\mathcal{P}_{\mathrm{vdw}} = \{(i,j) : i < j\}$ and corresponding lower-bound
distances $\ell_{ij}$ from $\mathcal{B}_{\mathrm{proj}}$. Candidate pairs are
restricted to valid, non-padding atoms whose chains are not joined by covalent
connectivity; single-atom ion chains are also excluded. For each remaining pair the
lower bound is defined from the vdW radii of the two atoms, relaxed by
a tolerance $\beta$,
\[
  \ell_{ij} = (1-\beta)\left(r_i^{\mathrm{vdw}} + r_j^{\mathrm{vdw}}\right).
\]
The buffer admits the mild contacts that occur in experimentally resolved
structures, so that only clear overlaps are treated as violations: at the value
we use, $\beta = 0.15$, two atoms must approach within $85\%$ of their summed van
der Waals radii before the pair is flagged. The projection therefore targets
inter-chain steric overlap.
 
 For each batch element we compute all
pairwise distances over the candidate set,
\[
  d_{ij} = \max\!\left(\|X'_i - X'_j\|_2,\ \varepsilon\right),
\]
where $\varepsilon$ is a small positive constant guarding division by zero, and
convert these into penetration depths,
\[
  p_{ij} = \max(\ell_{ij} - d_{ij},\ 0).
\]
Pairs with $p_{ij} > 0$ are considered violated. If no candidate pair violates
its lower bound, the batch element is left unchanged. To keep the projection minimal and stable we do not update all violated pairs.
Instead we select the top-$K$ pairs by penetration depth, where
$K = 200$. Each selected pair $(i,j)$ contributes the
damped, symmetric push-apart update
\[
  u_{ij} = \tfrac{1}{2}\,\alpha\, p_{ij}\, \hat{r}_{ij},
  \qquad
  \hat{r}_{ij} = \frac{X'_i - X'_j}{d_{ij}},
\]
which is added to atom $i$ and subtracted from atom $j$. The factor $\tfrac{1}{2}$
splits the correction evenly between the two atoms, and the damping factor
$\alpha$ sets how much of the remaining overlap a
single step removes: the pair separates by $\alpha\,p_{ij}$. For the Boltz-2 base model we use $\alpha = 0.25$, which removes a quarter of
the penetration per step and leaves the remainder to the denoiser; for
OpenFold-3 we use $\alpha = 1$. The under-relaxation in the Boltz-2 setting is
deliberate: the projection is applied at every step of a $40$-step window, so
a small repeated nudge is sufficient to reach feasibility while keeping any
individual correction far smaller than the structural changes the denoiser
itself makes.
 
The selected pair updates are accumulated into a per-atom displacement tensor $D$
before being applied. This ensures that if an atom participates in several violated pairs, its total
displacement is capped at $\delta_{\max}$ after summing all pairwise
corrections rather than each correction being clipped independently:
\[
  D_i \leftarrow D_i \min\!\left(
    1,\ \frac{\delta_{\max}}{\max(\|D_i\|_2,\ \varepsilon)}
  \right).
\]
The clipped displacement is added to the coordinates, returning
$X' = \hat{X} + D$. 
 
Our design makes the projection intentionally local, sparse, and minimal: local
because each update acts only on the atoms involved in a violated pair; sparse
because only the largest vdW overlaps are selected; and minimal because
each correction is damped by $\alpha$ and the accumulated displacement of each
atom is explicitly capped. The goal is therefore not to solve a full molecular
optimization problem or enforce every chemical validity constraint, but to remove
the most severe inter-chain steric conflicts while staying as close as possible
to the denoised structure predicted by the model. 

% The settings used for each base model are listed at the end of
% Section~\ref{sec:ldg_projection}; Table~\ref{tab:sensitivity} reports a
% sensitivity analysis over them.
\subsection{Ligand Distance-Geometry Projection}
\label{sec:ldg_projection}

\textsc{LDGProjection} (Algorithm~\ref{alg:ldg_projection}) corrects the
internal geometry of every ligand in the denoised estimate $\hat{X}$ toward the
distance-geometry, planarity and, optionally, chirality constraints implied by
its molecular graph. It is active only inside the late-step window
$t_{\tau} \le \texttt{projection\_start\_t}$, moves atoms by bounded,
per-atom-accumulated displacements, requires no additional network
evaluations, and leaves the denoiser and its weights untouched. Constraints are
built once per trajectory by \textsc{BuildLdgArgs}; the operator returns the
corrected coordinates together with a chirality latch $L$ that the sampler
carries from one step to the next.

% ---------------------------------------------------------------------------
% ALGORITHMS
% ---------------------------------------------------------------------------

\begin{algorithm}[t]
\caption{\textsc{LDGProjection}}
\label{alg:ldg_projection}
\begin{algorithmic}[1]
\Require Denoised coordinates $\hat{X} \in \mathbb{R}^{B \times N_{\mathrm{atom}} \times 3}$;
  batch features $\mathcal{B}_{\mathrm{proj}}$;
  normalised sampling time $t_{\tau} \in (0,1]$
\Require Buffers $\beta_{\mathrm{b}}, \beta_{\mathrm{a}}, \beta_{\mathrm{c}}, \beta_{\mathrm{p}}$,
  % damping $\alpha_{\mathrm{ldg}}$, 
  number of sweeps $N_{\mathrm{sw}}$,
  displacement cap $\delta_{\mathrm{ldg}}$;
  numerical floor $\varepsilon$
\Require Chirality latch $L \in \{0,1\}^{B \times |\mathcal{Q}|}$, zero at the start of
  each trajectory
  % flag $\texttt{force} \in \{0,1\}$,default $0$
\Ensure Projected coordinates $X'$; updated latch $L$
\If{$t_{\tau} >\texttt{projection\_start\_t}$  
% \textbf{and not} \texttt{force}
}
  \State \Return $(\hat{X}, L)$ \Comment{projection inactive outside the window}
\EndIf
\State $(\mathcal{P}_{\mathrm{BA}}, \mathcal{P}_{\mathrm{CL}}, \ell, h, \mathcal{G}, \mathcal{Q})
  \leftarrow \textsc{BuildLdgArgs}(\mathcal{B}_{\mathrm{proj}})$
  \Comment{cached}
\State $X' \leftarrow \mathrm{clone}(\hat{X})$
\For{$b = 1,\dots,B$}
  \State $C \leftarrow X'_b$

    \State $(C, L_b) \leftarrow \textsc{ChiralityStep}(C, L_b)$
      \Comment{Alg.~\ref{alg:ldg_chirality}}

  \For{$s = 1,\dots,N_{\mathrm{sw}}$}
    \State $C \leftarrow \textsc{IntervalSweep}(C, \mathcal{P}_{\mathrm{BA}})$
      \Comment{Alg.~\ref{alg:ldg_interval_sweep}: bonds, angles}
    \State $C \leftarrow \textsc{IntervalSweep}(C, \mathcal{P}_{\mathrm{CL}})$
      \Comment{Alg.~\ref{alg:ldg_interval_sweep}: internal clashes}
    \State $C \leftarrow \textsc{IntervalSweep}(C, \mathcal{P}_{\mathrm{BA}})$
      \Comment{Alg.~\ref{alg:ldg_interval_sweep}: re-seat bonds, angles}
    \State $C \leftarrow \textsc{PlanarSweep}(C, \mathcal{G})$
      \Comment{Alg.~\ref{alg:ldg_planar_sweep}: planarity}
  \EndFor
  \State $X'_b \leftarrow C$
\EndFor
\State \Return $(X', L)$
\end{algorithmic}
\end{algorithm}

\begin{algorithm}[t]
\caption{\textsc{IntervalSweep}}
\label{alg:ldg_interval_sweep}
\begin{algorithmic}[1]
\Require Coordinates $C \in \mathbb{R}^{N_{\mathrm{atom}} \times 3}$;
  pair set $\mathcal{R}$ with bounds $[\ell_{ij}, h_{ij}]$
\Ensure Updated coordinates
\State $D \leftarrow 0$ \Comment{per-atom displacement accumulator}
\ForAll{$(i,j) \in \mathcal{R}$} \Comment{every pair; no top-$K$}
  \State $d_{ij} \leftarrow \max(\|C_i - C_j\|_2,\ \varepsilon)$
  \State $\Delta_{ij} \leftarrow
    \begin{cases}
      \ell_{ij} - d_{ij} & d_{ij} < \ell_{ij} \\
      h_{ij} - d_{ij}    & d_{ij} > h_{ij} \\
      0                  & \text{otherwise}
    \end{cases}$
    \Comment{signed violation}
  \State $u_{ij} \leftarrow \tfrac{1}{2}\,
  % \alpha_{\ma thrm{ldg}}\
  \Delta_{ij}\, (C_i - C_j)/d_{ij}$
    \Comment{push apart or pull together}
  \State $D_i \leftarrow D_i + u_{ij}$;\quad $D_j \leftarrow D_j - u_{ij}$
\EndFor
\For{$i = 1,\dots,N_{\mathrm{atom}}$}
  \State $D_i \leftarrow D_i \cdot
    \min\!\big(1,\ \delta_{\mathrm{ldg}} / \max(\|D_i\|_2,\ \varepsilon)\big)$
    \Comment{per-atom displacement cap}
\EndFor
\State \Return $C + D$
\end{algorithmic}
\end{algorithm}

\begin{algorithm}[t]
\caption{\textsc{PlanarSweep}}
\label{alg:ldg_planar_sweep}
\begin{algorithmic}[1]
\Require Coordinates $C \in \mathbb{R}^{N_{\mathrm{atom}} \times 3}$;
  planar groups $\mathcal{G}$; planarity buffer $\beta_{\mathrm{p}}$
\Ensure Updated coordinates
\State $D \leftarrow 0$
\ForAll{$g \in \mathcal{G}$}
  \State $\mu_g \leftarrow \frac{1}{|g|} \sum_{k \in g} C_k$
  \State $\Sigma_g \leftarrow \sum_{k \in g} (C_k - \mu_g)(C_k - \mu_g)^{\top}$
  \State $n_g \leftarrow$ smallest-eigenvalue eigenvector of $\Sigma_g$
    \Comment{plane normal}
  \ForAll{$k \in g$}
    \State $\zeta_k \leftarrow (C_k - \mu_g)^{\top} n_g$
      \Comment{signed out-of-plane offset}
    \State $D_k \leftarrow D_k - \mathrm{sign}(\zeta_k)\,
      \max(|\zeta_k| - \beta_{\mathrm{p}},\ 0)\; n_g$
  \EndFor
\EndFor
\For{$i = 1,\dots,N_{\mathrm{atom}}$}
  \State $D_i \leftarrow D_i \cdot
    \min\!\big(1,\ \delta_{\mathrm{ldg}} / \max(\|D_i\|_2,\ \varepsilon)\big)$
    \Comment{per-atom displacement cap}
\EndFor
\State \Return $C + D$
\end{algorithmic}
\end{algorithm}

\begin{algorithm}[t]
\caption{\textsc{ChiralityStep}}
\label{alg:ldg_chirality}
\begin{algorithmic}[1]
\Require Coordinates $C \in \mathbb{R}^{N_{\mathrm{atom}} \times 3}$;
  latch $L \in \{0,1\}^{|\mathcal{Q}|}$; quadruples $\mathcal{Q}$ with
  handedness $\chi_q$, margin $v_q$ and signed volume $V_q$
  (~\eqref{eq:signed_volume})
\Ensure Updated coordinates and latch
\State $V_q \leftarrow V_q(C)$ for all $q \in \mathcal{Q}$
\State $\mathcal{A} \leftarrow \{q \in \mathcal{Q} : L_q = 1 \ \lor\ \chi_q V_q < 0\}$
  \Comment{inverted, or still latched}
\If{$\mathcal{A} = \emptyset$}
  \State \Return $(C, L)$ \Comment{correct centres are never edited}
\EndIf
\State $D \leftarrow 0$
\ForAll{$q \in \mathcal{A}$ with $\chi_q V_q < v_q$}
  \State $\lambda_q \leftarrow  (\chi_q v_q - V_q) \,/\,
    \max(\|\nabla V_q\|_F^2,\ \varepsilon)$
    \Comment{Newton step}
  \For{$k = 0,\dots,3$}
    \State $D_{q_k} \leftarrow D_{q_k} + \lambda_q\, \nabla_{q_k} V_q$
  \EndFor
\EndFor
\For{$i = 1,\dots,N_{\mathrm{atom}}$}
  \State $D_i \leftarrow D_i \cdot
    \min\!\big(1,\ \delta_{\mathrm{ldg}} / \max(\|D_i\|_2,\ \varepsilon)\big)$
    \Comment{per-atom displacement cap}
\EndFor
\State $C \leftarrow C + D$
\State $L_q \leftarrow
  \mathbf{1}\big[q \in \mathcal{A} \land \chi_q V_q(C) < v_q\big]\ \ \forall q \in \mathcal{Q}$
\State \Return $(C, L)$
\end{algorithmic}
\end{algorithm}

\paragraph{Constraint construction.}
\textsc{BuildLdgArgs} (Algorithm~\ref{alg:ldg_projection}) builds the constraints
once per trajectory from the ligand molecular graphs in $\mathcal{B}_{\mathrm{proj}}$.
For every ligand with at least two heavy atoms, the
candidate set $\mathcal{P}$ contains all heavy-atom pairs $(i,j)$, $i<j$, with
RDKit distance-geometry bounds $[\ell^{\mathrm{rd}}_{ij}, h^{\mathrm{rd}}_{ij}]$.
Bonded (1--2) pairs $\mathcal{P}_{\mathrm{bond}}$ and angle (1--3) pairs
$\mathcal{P}_{\mathrm{ang}}$ form
$\mathcal{P}_{\mathrm{BA}} = \mathcal{P}_{\mathrm{bond}} \cup \mathcal{P}_{\mathrm{ang}}$,
and all remaining pairs, those that are neither bonded nor separated by one
atom, form the clash set $\mathcal{P}_{\mathrm{CL}}$. The bounds are widened by the buffers,
\[
  [\ell_{ij}, h_{ij}] = \big[(1-\beta)\,\ell^{\mathrm{rd}}_{ij},\ (1+\beta)\,h^{\mathrm{rd}}_{ij}\big],
  \quad
  \beta =
  \begin{cases}
    \beta_{\mathrm{b}} & \text{1--2 pairs only} \\
    \beta_{\mathrm{a}} & \text{1--3 pairs only} \\
    \min(\beta_{\mathrm{b}}, \beta_{\mathrm{a}}) & \text{both}
  \end{cases}
\]
for $(i,j) \in \mathcal{P}_{\mathrm{BA}}$, while pairs in $\mathcal{P}_{\mathrm{CL}}$
receive only a lower bound, $[\ell_{ij}, h_{ij}] = [(1-\beta_{\mathrm{c}})\,\ell^{\mathrm{rd}}_{ij},\ +\infty)$. The planar groups $\mathcal{G}$ are the aromatic five- and
six-membered rings and the planar $\mathrm{C{=}C}$ groups (the double bond with its
four substituents). $\mathcal{Q}$ holds the assigned
sp$^3$ stereocentres with at least three heavy neighbours, each as a quadruple
$q = (q_0, \dots, q_3)$ of the centre and its three heavy neighbours, whose handedness
is the sign of the signed volume
\begin{equation}
  V_q(C) = (C_{q_1} - C_{q_0}) \cdot \big[(C_{q_2} - C_{q_0}) \times (C_{q_3} - C_{q_0})\big].
  \label{eq:signed_volume}
\end{equation}
The target handedness $\chi_q = \mathrm{sign}\,V_q(X^{\mathrm{ref}})$ and the margin
$v_q = \kappa\,|V_q(X^{\mathrm{ref}})|$ are read from the RDKit-embedded reference
conformer $X^{\mathrm{ref}}$.
% ^ if the code picks a specific three of four neighbours (e.g. by CIP priority
%   or atom index), add the rule here.

% ---------------------------------------------------------------------------
% EXPLANATORY PARAGRAPHS
% ---------------------------------------------------------------------------

\paragraph{Interval constraints.}
Every heavy-atom pair carries an admissible distance interval: bonded and angle
pairs are held to a two-sided interval around their RDKit bounds, all other
pairs to a lower bound only, each widened by a class-specific tolerance so that
only clear violations are corrected. \textsc{IntervalSweep}
(Algorithm~\ref{alg:ldg_interval_sweep}) visits every pair in a set, computes
the signed violation $\Delta_{ij}$, positive when the pair is too close,
negative when too far, and applies the symmetric update
$u_{ij} = \tfrac12\Delta_{ij}\,\hat r_{ij}$ to atom $i$
and $-u_{ij}$ to atom $j$, pushing the pair apart or pulling it together as
required. All updates in a pass are accumulated into a per-atom tensor $D$, and
each atom's total displacement is capped at $\delta_{\mathrm{ldg}}$ before it
is applied, so an atom in many violated pairs is bounded on its net motion
rather than on each contribution.

\paragraph{Sweep order.}
Each of the $N_{\mathrm{sw}}$ sweeps runs the interval pass on
$\mathcal{P}_{\mathrm{BA}}$, then on $\mathcal{P}_{\mathrm{CL}}$, then on
$\mathcal{P}_{\mathrm{BA}}$ again, and finally \textsc{PlanarSweep}. Resolving
an internal clash can stretch the bonds adjacent to it, so bonds and angles are
re-seated after the clash pass; planarity comes last because it moves whole
ring systems that the pairwise passes would otherwise disturb.
\textsc{PlanarSweep} (Algorithm~\ref{alg:ldg_planar_sweep}) fits a plane to
each aromatic ring or $\mathrm{C{=}C}$ group by taking the smallest-eigenvalue
eigenvector of the group's coordinate covariance as its normal $n_g$, and moves
each member atom toward the plane by its
out-of-plane offset in excess of $\beta_{\mathrm{p}}$. The sign of $n_g$ is
arbitrary, but the update depends only on $\mathrm{sign}(\zeta_k)\,n_g$, which
is not. Because the cap is applied within each pass, an atom may move by up to
$\delta_{\mathrm{ldg}}$ per pass over the course of one diffusion step.

\paragraph{Chirality.}
\textsc{ChiralityStep} (Algorithm~\ref{alg:ldg_chirality}) runs
once per call before the sweeps. Each stereocentre in $\mathcal{Q}$ is
represented by the signed volume $V_q$ of Eq.~\eqref{eq:signed_volume}, whose
sign gives its handedness. A centre whose sign disagrees with the target
$\chi_q$ receives a damped Newton step on $V_q$ toward $\chi_q v_q$, distributed
over its four atoms along $\nabla V_q$. Because $V_q$ passes through zero as a
centre inverts, a correction that stopped at the first sign change would leave
a nearly planar centre that the denoiser could flip back; the latch $L_q$
therefore keeps a centre active until $\chi_q V_q$ exceeds the margin $v_q$.
The latch is initialised to zero once per trajectory and threaded through every
call, so a centre latched at one step remains active at the next. Correct,
unlatched centres are never edited.

% \paragraph{Cost.}
% Constraint sets are built once per trajectory and cached. Each pass is a gather
% over a fixed index set followed by a scatter-add into $D$, and each plane fit is
% a $3{\times}3$ eigendecomposition; no additional network evaluations are
% required.

% ---------------------------------------------------------------------------
% SAMPLER CHANGES (appendix, projected sample-diffusion) -- as before
% ---------------------------------------------------------------------------
% Before the diffusion loop:
%   \State $L \leftarrow 0$ \Comment{chirality latch, once per trajectory}
% In each step, on the denoised estimate, after VDWProjection:
%   \State $(\hat{X}, L) \leftarrow \textsc{LDGProjection}(\hat{X}, \mathcal{B}_{\mathrm{proj}}, t_n, L)$
% After the final Euler step, if N_post > 0:
% \For{$r = 1,\dots,N_{\mathrm{post}}$}
%   \State $(X, L) \leftarrow \textsc{LDGProjection}(X, \mathcal{B}_{\mathrm{proj}}, \cdot, L, \texttt{force}{=}1)$
%     \Comment{time gate bypassed}
% \EndFor

% ---------------------------------------------------------------------------
% FINAL PARAMETERS (as supplied; sloppy group keeps the identifiers in-margin)
% ---------------------------------------------------------------------------

\begingroup\sloppy
\paragraph{Ours (Boltz).}
Unless stated otherwise the vdW projection uses
\mbox{$\texttt{projection\_start\_t} = 0.2$},
$\alpha = 0.25$,
$\delta_{\max} = 0.45$\,\AA{},
$K = 200$,
$\beta = 0.15$,
$\varepsilon = 10^{-8}$,
and one sweep per diffusion step.
Unless stated otherwise the LDG projection uses
$\delta_{\mathrm{ldg}} = 0.2$\,\AA{},
$\beta_{\mathrm{b}} = 0.24$,
$\beta_{\mathrm{a}} = 0.24$,
$\beta_{\mathrm{c}} = 0.19$,
$\beta_{\mathrm{p}} = 0.19$\,\AA{},
$N_{\mathrm{sw}} = 3$,
% $N_{\mathrm{post}} = 0$.
\paragraph{Ours (OpenFold).}
Unless stated otherwise the vdW projection uses
$\alpha = 1.0$,
$\delta_{\max} = 0.3$\,\AA{},
$K = 200$,
$\beta = 0.15$,
$\varepsilon = 10^{-8}$,
and one sweep per diffusion step.
Unless stated otherwise the LDG projection uses
$\delta_{\mathrm{ldg}} = 0.25$\,\AA{},
$\beta_{\mathrm{b}} = 0.24$,
$\beta_{\mathrm{a}} = 0.10$,
$\beta_{\mathrm{c}} = 0.19$,
$\beta_{\mathrm{p}} = 0.20$\,\AA{},
$\kappa = 0.1$,
$N_{\mathrm{sw}} = 3$
% $N_{\mathrm{post}} = 10$,
and the same $\texttt{projection\_start\_t}$ as
Algorithm~\ref{alg:vdw_projection}.
\par\endgroup
\section{Sampler dynamics and the projection window}
\label{app:sampler_dynamics}

To understand where in the reverse process a correction can take effect, we
record the iterate $x_t$ and the denoised estimate $\hat{x}_0$ at every step
of an unprojected Boltz-2 run (200 steps) on four CASP16 targets (M1212, M1296,
T1234, T1249v2) and compare each against the final structure. Churn is active
up to step 154 and off thereafter (shaded region in all figures).

\begin{figure}[t]
  \centering
  \includegraphics[width=\linewidth]{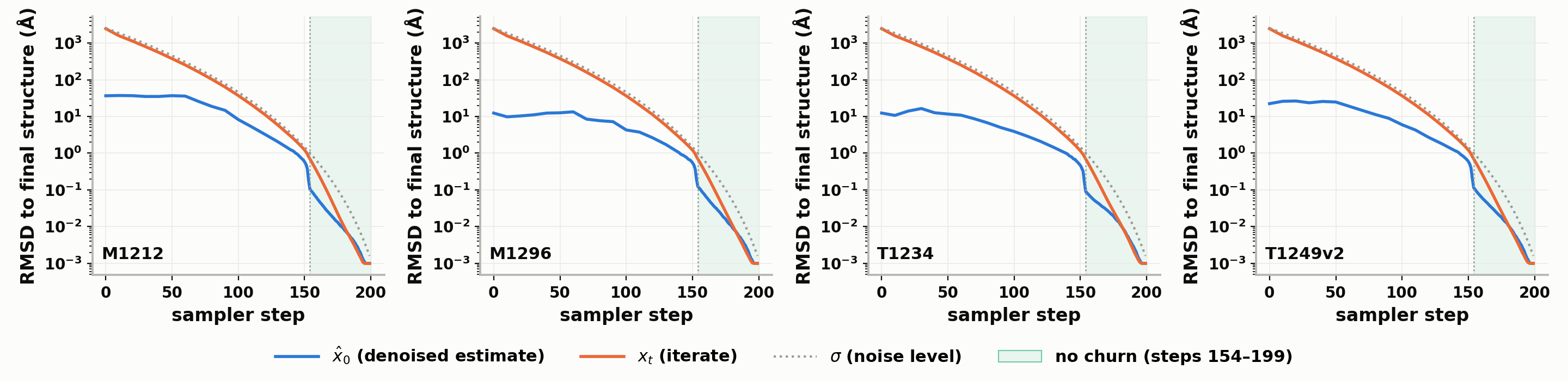}
  \caption{RMSD to the final structure of the denoised estimate $\hat{x}_0$
  and the iterate $x_t$ at each sampler step (log scale), with the noise level
  $\sigma$ for reference. $x_t$ tracks $\sigma$ until churn switches off;
  $\hat{x}_0$ drops by an order of magnitude at the churn boundary. Shaded:
  steps without churn.}
  \label{fig:traj_rmsd}
\end{figure}

\begin{figure}[t]
  \centering
  \includegraphics[width=\linewidth]{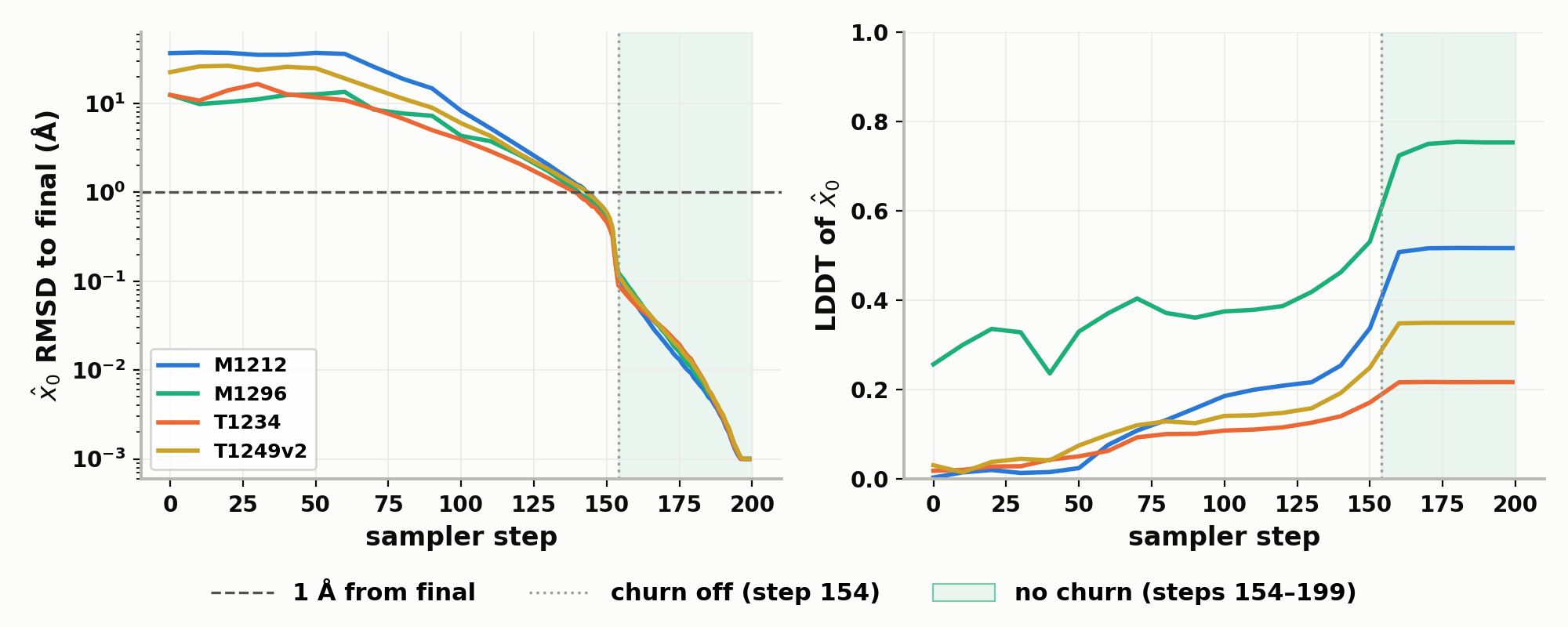}
  \caption{Summary across the four targets. Left: RMSD of $\hat{x}_0$ to the
  final structure; all targets cross $1$\,\AA{} (dashed) just before the churn
  boundary and fall to $\sim$$0.1$\,\AA{} immediately after. Right: LDDT of
  $\hat{x}_0$, which levels off within a few steps of the boundary.}
  \label{fig:traj_summary}
\end{figure}

\paragraph{The structure is fixed when churn ends.}
The iterate $x_t$ is dominated by noise for most of sampling: its RMSD to the
final structure tracks the noise level $\sigma$ almost exactly until the churn
boundary (Fig.~\ref{fig:traj_rmsd}). The denoised estimate behaves
differently. It starts 10--40\,\AA{} from the final structure, approaches it
steadily, and then drops by an order of magnitude, from $\sim$$1$\,\AA{} to
$\sim$$0.1$\,\AA{}, at the step where churn switches off
(Fig.~\ref{fig:traj_summary}, left). The all-atom LDDT of $\hat{x}_0$ levels
off within a few steps of the boundary (Fig.~\ref{fig:traj_summary}, right).
The remaining $\sim$45 steps refine the structure by less than $0.1$\,\AA{}.

\begin{figure}[t]
  \centering
  \includegraphics[width=\linewidth]{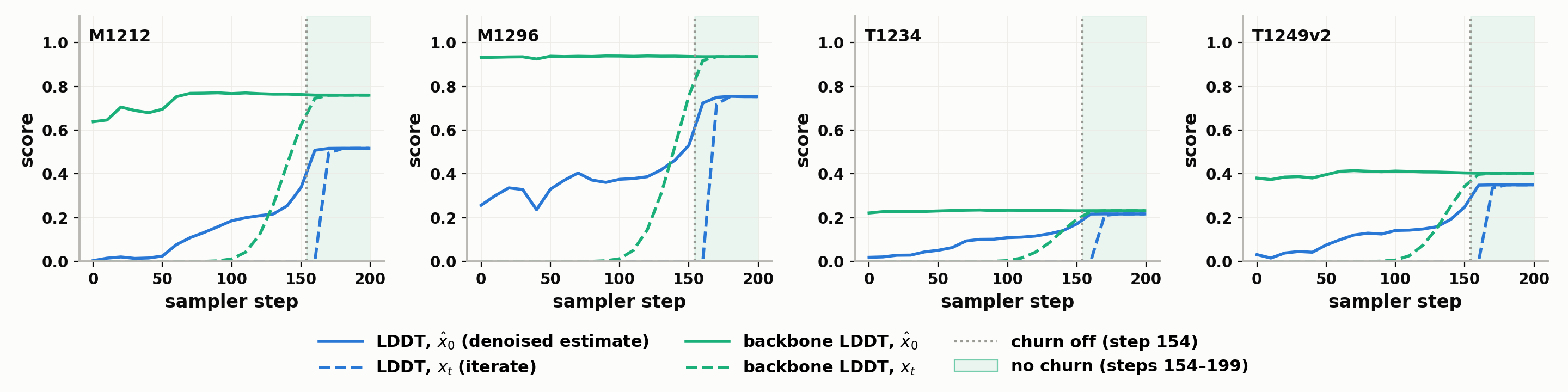}
  \caption{All-atom and backbone LDDT of $\hat{x}_0$ (solid) and $x_t$
  (dashed) at each sampler step. The backbone of $\hat{x}_0$ is close to its
  final accuracy from the first step; $x_t$ only becomes a meaningful
  structure after churn switches off.}
  \label{fig:traj_lddt}
\end{figure}

\paragraph{Why project $\hat{x}_0$.}
Van der Waals and ligand-geometry constraints are defined on clean structures.
The denoised estimate is a clean structure from the first step: its backbone
LDDT is at or near its final value from step~0, and only the all-atom LDDT
improves over sampling (Fig.~\ref{fig:traj_lddt}). The iterate $x_t$, by
contrast, has near-zero all-atom LDDT until $\sim$step 160, and its backbone
LDDT only rises between steps 100 and 160. Projecting $\hat{x}_0$ therefore
applies the constraints to the model's current best estimate of the structure
rather than to noise, and the correction reaches the trajectory through the
drift (line~\ref{alg:xnoisy}). Once churn is off, $x_t$ and $\hat{x}_0$ are
within $\sigma \le 1$\,\AA{} of each other, and in this regime projecting
either one gives the same validity and accuracy
(Table~\ref{tab:sen_position}).

\begin{figure}[t]
  \centering
  \includegraphics[width=\linewidth]{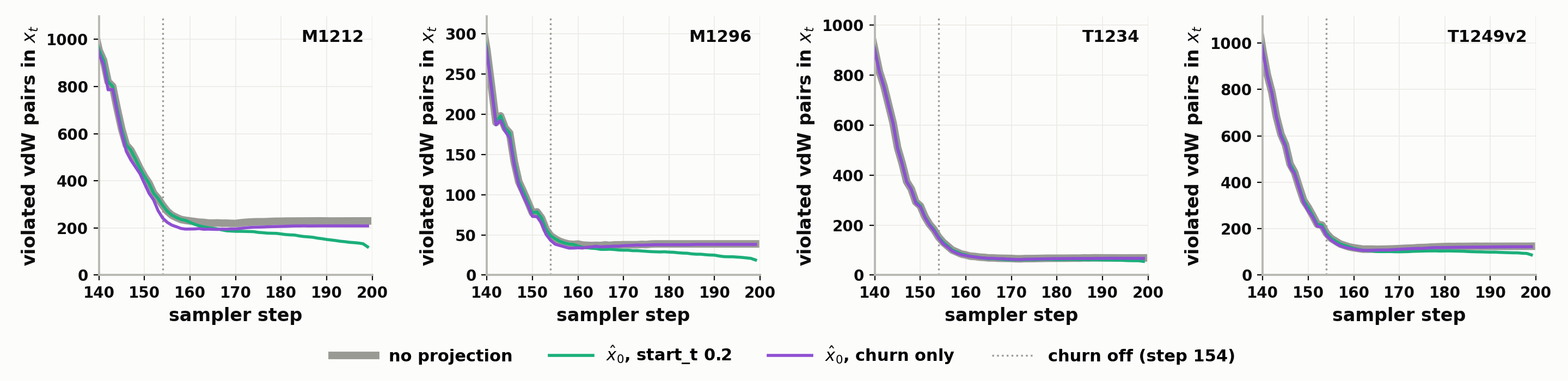}
  \caption{Number of violated van der Waals pairs in $x_t$ over steps
  140--199. Projecting in the tail
  ($\texttt{projection\_start\_t}=0.2$, green) reduces violations steadily to the
  end of sampling.}
  \label{fig:vdw_violations}
\end{figure}

\paragraph{Why project only after churn.}
Figure~\ref{fig:vdw_violations} tracks the number of violated van der Waals
pairs in $x_t$ over the last 60 steps. Projecting only during churn has no
visible effect: at no point does the count fall below the unprojected run,
before or after the churn boundary. Each correction is at most
$\delta_{\max} = 0.45$\,\AA{} per atom, whereas each churn step injects noise
with per-coordinate standard deviation of at least $1.5$\,\AA{}, so the
correction is overwritten before it can accumulate. With
$\texttt{projection\_start\_t} = 0.2$, the count separates from the
unprojected run at $\sim$step 160 and falls steadily until the end of
sampling. In this tail no fresh noise is injected, and the denoiser itself
changes $\hat{x}_0$ by less than $0.1$\,\AA{} RMSD over all remaining steps
(Fig.~\ref{fig:traj_summary}). Successive bounded corrections are therefore
retained and add up. The benchmark results agree: churn-only projection leaves
physical validity at $0.02$ ($\hat{x}_0$) and $0.04$ ($x_t$), while tail
projection reaches $1.00$ with unchanged accuracy (Table~\ref{tab:sen_position}).

The counts in Figure~\ref{fig:vdw_violations} do not reach zero violations because the diagnostic
is stricter than the evaluation: it is the projection's own criterion. A pair counts
as violated if its two atoms lie in different chains (excluding covalently linked
chains and single-atom ions) and are closer than $(1-\beta)(r_i + r_j)$ with
$\beta = 0.15$, i.e.\ $0.85\,(r_i + r_j)$, with no further tolerance. Physical
validity in our benchmarks follows the Boltz definition \citep{boltz2}, whose
steric-clash test flags a pair of chains as clashing if any two of their atoms are
closer than $0.75\,(r_i + r_j)$, using the same van der Waals radii. Almost all pairs
remaining in the figure are mild contacts between these two thresholds, inside the
spacing the projection aims for but outside what the Boltz criterion treats as a
clash. At the last step of the $\hat{x}_0$ run with \texttt{projection\_start\_t}~$=0.2$,
1{,}401 of the 1{,}402 remaining pairs (four targets, five samples each) lie between
$0.75$ and $0.85\,(r_i + r_j)$, and 19 of the 20 structures are clash-free. Without
projection, 1{,}037 of 2{,}309 pairs fall below $0.75\,(r_i + r_j)$ and no structure
is clash-free. The figure is meant to show when corrections are retained, not to
measure validity.


\subsection{Projection Dynamics}
\label{app:diffusion_projection_diagnostics}
\begin{figure}[t]
    \centering

    \begin{subfigure}[t]{0.48\linewidth}
        \centering
        \includegraphics[width=\linewidth]{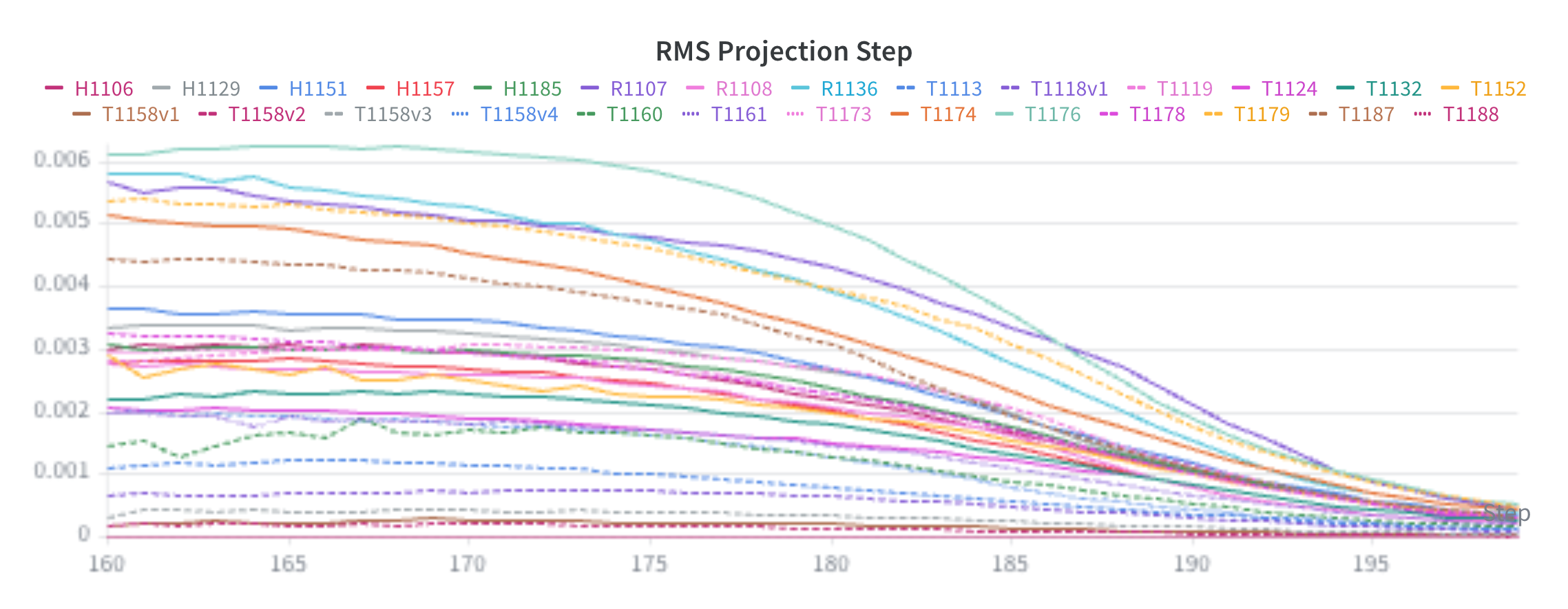}
        \caption{RMS projection step.}
        \label{fig:rms_projection_step}
    \end{subfigure}
    \hfill
    \begin{subfigure}[t]{0.48\linewidth}
        \centering
        \includegraphics[width=\linewidth]{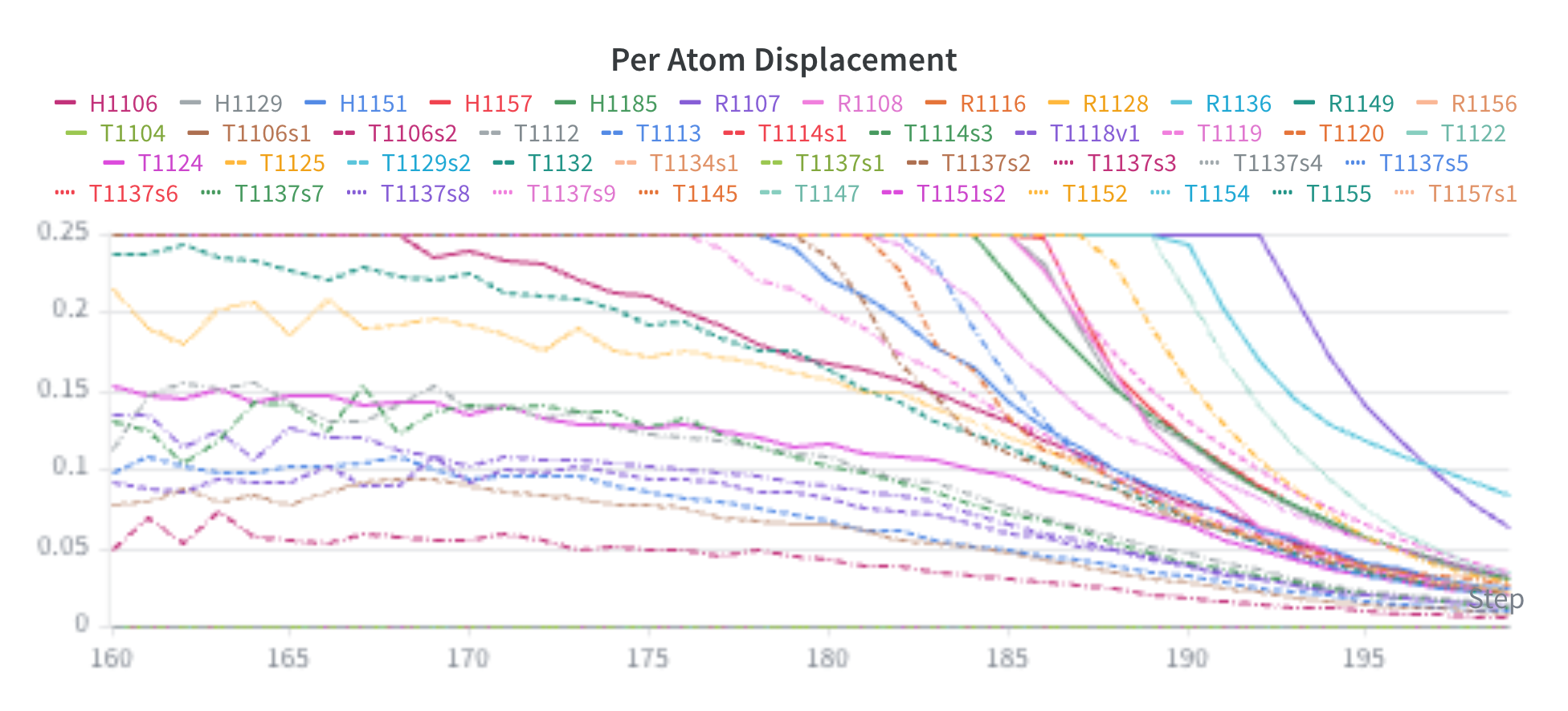}
        \caption{Maximum per-atom projection displacement.}
        \label{fig:per_atom_displacement}
    \end{subfigure}

    \vspace{0.75em}

  \begin{subfigure}[t]{0.8\linewidth}
    \centering
    \includegraphics[width=\linewidth]{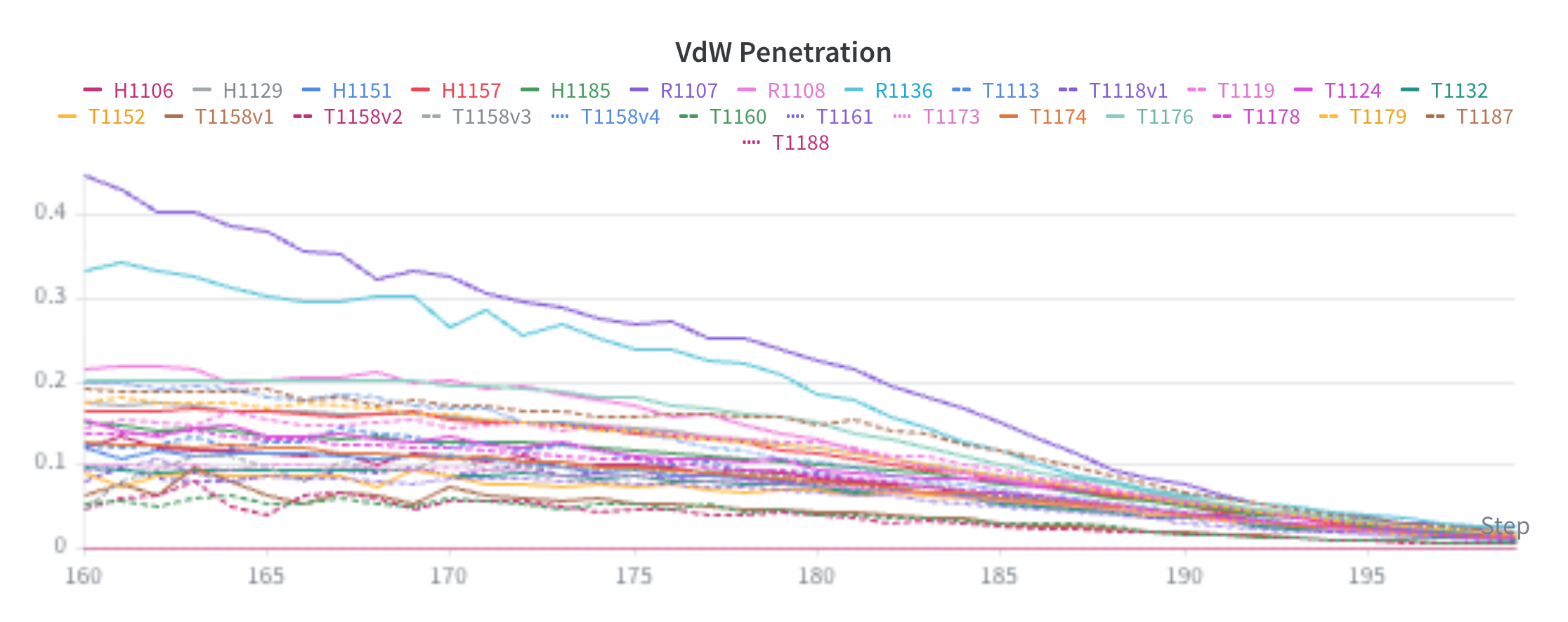}
    \caption{Mean VdW penetration depth.}
    \label{fig:vdw_penetration}
\end{subfigure}

    \caption{
    Each line corresponds to a different system in CASP15, with colors and line styles as labeled.
    }
    \label{fig:projection_late_traj}
\end{figure}

We posit that minimal, local coordinate updates can remove a large fraction of
physical-validity violations while remaining compute- and memory-efficient.
Figure~\ref{fig:projection_late_traj} illustrates the vdW projection dynamics on
CASP15 targets. Projection is activated only in the late refinement regime; in
this experiment, \texttt{projection\_start\_t}$=0.2$, corresponding to the final
$40$ steps of the standard $200$-step AF3-style sampler. We therefore plot only
steps $160$--$200$, since all projection quantities are zero before activation.
The observed dynamics show that the correction remains small and
displacement-limited while VDW penetration decreases over the final denoising
steps.

\textbf{RMS projection step} is the Root Mean Square (RMS) of the final capped per-atom displacement, $ D_i$, produced by
Algorithm~\ref{alg:vdw_projection}. The plotted quantity is
\[
\mathrm{RMS}_{\mathrm{proj}}
=
\sqrt{
\frac{1}{N_{\mathrm{atom}}}
\sum_{i=1}^{N_{\mathrm{atom}}}
\|\mathbf D_i\|_2^2
}.
\]
This value decreases throughout the trajectory, indicating that the
projection requires progressively smaller corrections as the structure becomes
more physically consistent.

% projection is a local repair operator % As our projection method operates through small, localized coordinate changes,
\textbf{Maximum per-atom displacement} shows the largest single-atom movement given by
\[
    d_{\max}
    =
    \max_i \|\mathbf D_i\|_2 .
\]

We cap the maximum per-atom displacement by the parameter \emph{projection\_max\_atom\_disp} in
Algorithm~\ref{alg:vdw_projection}. Although some atoms reach this cap in the
early steps, the maximum displacement rapidly decreases toward the end
of sampling, indicating that the correction becomes increasingly minimal as violations disappear. 
   
\textbf{VDW penetration depth} shows the mean van der Waals penetration depth over violated pairs
\[
    p_{ij}
    =
    \max\!\left(\ell_{ij} - \|\mathbf{x}_i-\mathbf{x}_j\|_2,\,0\right),
\]
where $\ell_{ij}$ is the VDW lower-bound distance. This quantity is nonzero only when two atoms are too close, i.e. when they clash. Therefore, lower penetration directly means weaker or fewer steric overlaps, and \(p_{ij}=0\) means that the pair satisfies the van der Waals separation constraint. Across CASP15 targets, the mean penetration consistently decreases toward zero, showing that
steric overlaps are progressively reduced as sampling approaches the terminal
structure. 

Interpreted together with
Figures~\ref{fig:rms_projection_step}--\ref{fig:per_atom_displacement}, this
reduction happens with small corrections: the RMS projection step is small,
the per-atom displacement is capped at
$\delta_{\max}$, and both shrink as the clashes are
resolved. This fits the dynamics described in
Appendix~\ref{app:sampler_dynamics}. By the time projection is activated,
the denoiser has settled the fold: $\hat{x}_0$ is within
$\sim$$0.1$\,\AA{} of the final structure, and its LDDT has stopped
changing. What remains are local steric clashes that the model does not
remove on its own during the remaining steps. Projection targets exactly these
clashes, so it acts as a sparse local repair rather than a global energy
minimization. The resulting corrections are cheap and stable, and they remove
the remaining clashes without rearranging the structure the model has already
settled on.

\section{Experimental Setup}
\label{app:experimental_setup}

\subsection{Evaluation Criteria}

We evaluated all methods on the same target sets, using identical MSA settings,
random seed (42), and the same post-processing and metric scripts.
For each target, we ran inference for 5 samples, with the default sampling settings for Boltz-2 and OpenFold-3,
unless otherwise stated. Our projection method uses the same
weights and sampling schedule, with only the additional
inference-time projection enabled. 

We measured structural accuracy with OpenStructure using the same evaluation
pipeline as Boltz-1. We computed physical validity using the Boltz
physical-validity checks, which include ligand bond lengths, bond angles,
internal ligand clashes, chirality, E/Z stereochemistry, planarity, and
inter-chain steric clashes. A prediction was counted as physically valid only if it passes all considered violation checks. We used the same validity definition as Boltz, but
corrected a bug in the planarity check; as a result, Boltz validity values
differ (in the positive direction) from those reported in the original Boltz paper. 

We measured runtime as wall-clock inference time per target, including diffusion
sampling and excluding pre-processing steps like MSA generation. We evaluated GPU memory with PyTorch CUDA peak-memory counters. All runtime and memory comparisons were run on A100 80GB GPUs under the same software environment. For each method and dataset,
we reported mean metrics across targets; uncertainty is computed by bootstrap
resampling over targets.

We have used the following metrics to evaluate our method:
\begin{itemize}
\label{metrics}
    \item \textbf{LDDT} (\(\uparrow\)): Local Distance Difference Test, measuring local structural accuracy by comparing predicted and reference inter-atomic distances. Higher values indicate better local geometry.

    \item \textbf{DockQ} (\(\uparrow\)): Interface quality score for protein--protein or biomolecular docking interfaces, combining interface RMSD, ligand RMSD, and native contact recovery. Higher values indicate more accurate interface placement.

    \item \textbf{LDDT-PLI} (\(\uparrow\)): Protein--ligand interaction score, measuring the accuracy of predicted protein--ligand contacts and binding-site geometry. Higher values indicate better recovery of ligand interactions.

    \item \textbf{LRMSD\(<2\)} (\(\uparrow\)): Fraction of protein--ligand predictions with ligand RMSD below \(2\,\text{\AA}\), typically after pocket or receptor alignment. Higher values indicate a larger fraction of near-native ligand poses.

    \item \textbf{Valid} (\(\uparrow\)): Fraction of predictions passing the physical validity checks, including steric clash, ligand distance-geometry, stereochemistry, and planarity criteria. Higher values indicate fewer physical violations.
\end{itemize}

Per-metric target counts are given in the $n$ row of
Table~\ref{tab:main_results}. DockQ requires a scored interface and
LDDT-PLI/LRMSD$_{<2}$ a scored ligand, so those columns rest on subsets.

\subsection{Benchmark targets}\label{sec:benchmark-targets}
Every method was run on the same query set per dataset: CASP15 (66), CASP16 (55), PoseBusters Monomer (134), PoseBusters Complex (169), Test (542), 966 targets in total. CASP targets are listed by their CASP target ID (subunit-level queries keep their suffix, e.g.\ s1), all other targets by PDB ID. Targets in \textcolor{red}{red} are excluded from Table~\ref{tab:main_results} because Boltz-2x runs out of memory on them on an 80\,GB A100 (CASP16: H1236, R1283v2, R1285; PoseBusters Complex: 7B2C, 7CNQ, 7SUC, 8AQL, 8G0V; Test: 8SRO). Boltz-2 and Ours (Boltz) complete all of them, and all comparisons in the table use the targets every Boltz method completed.

\paragraph{CASP15 (66 targets).}
{\small\raggedright H1106, H1129, H1151, H1157, H1185, R1107, R1108, R1116, R1117, R1117v2, R1128, R1136, R1149, R1156, T1104, T1106s1, T1106s2, T1112, T1113, T1114s1, T1114s3, T1118v1, T1119, T1120, T1122, T1124, T1125, T1129s2, T1132, T1134s1, T1137s1, T1137s2, T1137s3, T1137s4, T1137s5, T1137s6, T1137s7, T1137s8, T1137s9, T1145, T1147, T1151s2, T1152, T1154, T1155, T1157s1, T1157s2, T1158, T1158v1, T1158v2, T1158v3, T1158v4, T1160, T1161, T1173, T1174, T1176, T1178, T1179, T1183, T1185s1, T1185s2, T1185s4, T1187, T1188, T1194\par}

\paragraph{CASP16 (55 targets (52 of 55 in Table~\ref{tab:main_results})).}
{\small\raggedright H1202, H1204, H1213, H1222, H1223, H1225, H1232, H1233, \textcolor{red}{H1236}, M1209, M1211, M1212, M1296, R1203, R1205, R1209, R1212, R1242, R1260, R1261, R1261v1, R1262, R1262v1, R1263, R1263v1, R1264, R1264v1, R1283v1, \textcolor{red}{R1283v2}, \textcolor{red}{R1285}, R1286, R1296, T1201, T1206, T1210, T1212, T1214, T1220s1, T1227s1, T1234, T1235, T1249v1, T1249v2, T1266, T1269, T1272s2, T1272s3, T1272s4, T1272s5, T1272s6, T1272s7, T1272s8, T1272s9, T1280, T1299\par}

\paragraph{PoseBusters Monomer (134 targets).}
{\small\raggedright 5SAK, 5SB2, 5SD5, 5SIS, 6M2B, 6M73, 6T88, 6TW7, 6WTN, 6XBO, 6XCT, 6XG5, 6XM9, 6YMS, 6YQW, 6YRV, 6Z0R, 6Z14, 6Z1C, 6ZAE, 6ZCY, 6ZPB, 7A9E, 7AKL, 7BCP, 7BJJ, 7BTT, 7CL8, 7CTM, 7D5C, 7DKT, 7ELT, 7EPV, 7ES1, 7F51, 7F5D, 7F8T, 7JG0, 7JMV, 7JXX, 7L5F, 7LCU, 7LOE, 7LT0, 7LZD, 7MGT, 7MMH, 7MWU, 7MY1, 7N03, 7N4W, 7N6F, 7N7H, 7NF3, 7NGW, 7NP6, 7NPL, 7NR8, 7NU0, 7O1T, 7OEO, 7OFF, 7OMX, 7OSO, 7OZC, 7P4C, 7P5T, 7PGX, 7PK0, 7PL1, 7PRM, 7Q27, 7Q2B, 7Q5I, 7QF4, 7QGP, 7QPP, 7R59, 7R7R, 7RC3, 7RH3, 7RKW, 7ROU, 7RWS, 7SDD, 7TOM, 7TUO, 7U3J, 7UQ3, 7USH, 7UYB, 7V14, 7V3N, 7V3S, 7V43, 7VBU, 7VC5, 7W05, 7W06, 7WCF, 7WDT, 7WJB, 7WPW, 7X9K, 7XBV, 7XFA, 7XI7, 7XQZ, 7YZU, 7Z1Q, 7ZF0, 7ZHP, 7ZOC, 7ZU2, 7ZXV, 8A1H, 8A2D, 8AAU, 8AEM, 8AP0, 8BOM, 8BTI, 8CNH, 8D39, 8DHG, 8DP2, 8EX2, 8EXL, 8EYE, 8FLV, 8G6P, 8GFD, 8HFN, 8HO0\par}

\paragraph{PoseBusters Complex (169 targets (164 of 169 in Table~\ref{tab:main_results})).}
{\small\raggedright 6TW5, 6VTA, 6XHT, 6YJA, 6YQV, 6YR2, 6YSP, 6YT6, 6YYO, 6Z2C, 6Z4N, 6ZC3, 6ZK5, 7A1P, 7A9H, 7AFX, 7AN5, \textcolor{red}{7B2C}, 7B94, 7BKA, 7BMI, 7BNH, 7C0U, 7C3U, 7C8Q, 7CD9, 7CIJ, \textcolor{red}{7CNQ}, 7CNS, 7CUO, 7DQL, 7DUA, 7E4L, 7EBG, 7ECR, 7ED2, 7FB7, 7FHA, 7FT9, 7JHQ, 7JY3, 7K0V, 7KB1, 7KC5, 7KM8, 7KQU, 7KRU, 7KZ9, 7L00, 7L03, 7L7C, 7LEV, 7LJN, 7LMO, 7LOU, 7M3H, 7M6K, 7MFP, 7MGY, 7MOI, 7MSR, 7MWN, 7MYU, 7N4N, 7N7B, 7NF0, 7NFB, 7NLV, 7NSW, 7NUT, 7NXO, 7O0N, 7ODY, 7OFK, 7OLI, 7OPG, 7OZ9, 7P1F, 7P1M, 7P2I, 7PIH, 7PJQ, 7POM, 7PRI, 7PT3, 7PUV, 7Q25, 7QE4, 7QFM, 7QHG, 7QHL, 7QTA, 7R3D, 7R6J, 7R9N, 7RNI, 7ROR, 7RSV, 7RZL, 7SCW, 7SFO, 7SIU, \textcolor{red}{7SUC}, 7SZA, 7T0D, 7T1D, 7T3E, 7TB0, 7TBU, 7TE8, 7TH4, 7THI, 7TM6, 7TS6, 7TSF, 7TXK, 7TYP, 7U0U, 7UAS, 7UAW, 7UJ4, 7UJ5, 7UJF, 7ULC, 7UMW, 7UTW, 7UXS, 7UY4, 7VB8, 7VKZ, 7VQ9, 7VWF, 7VYJ, 7WKL, 7WL4, 7WQQ, 7WUX, 7WUY, 7WY1, 7X5N, 7XG5, 7XJN, 7XPO, 7XRL, 7Z2O, 7Z7F, 7ZCC, 7ZDY, 7ZL5, 7ZTL, 7ZZW, 8AIE, \textcolor{red}{8AQL}, 8AUH, 8AY3, 8B8H, 8C3N, 8C5M, 8CSD, 8D19, 8D5D, 8DKO, 8DSC, 8EAB, 8F8E, 8FAV, 8FO5, \textcolor{red}{8G0V}, 8SLG\par}

\paragraph{Test (542 targets (541 of 542 in Table~\ref{tab:main_results})).}
{\small\raggedright 7FCT, 7QA1, 7QNL, 7QPK, 7QPL, 7QQD, 7QRR, 7QSJ, 7R1N, 7R5I, 7S2J, 7SC4, 7SFY, 7SPQ, 7T4W, 7TL6, 7TPU, 7TZL, 7U08, 7U71, 7UBA, 7UDL, 7UG2, 7UI4, 7UI8, 7UK8, 7UNH, 7UOI, 7UQU, 7UR2, 7URP, 7UWV, 7UYX, 7V34, 7VS6, 7VXT, 7W91, 7WJU, 7WKP, 7WMY, 7WNH, 7WOI, 7WRN, 7WUY, 7WWC, 7X2X, 7X6G, 7XBJ, 7XCA, 7XCC, 7XF1, 7XF2, 7XFS, 7XFU, 7XGS, 7XGT, 7XGV, 7XH4, 7XHZ, 7XL7, 7XMJ, 7XMW, 7XNJ, 7XP0, 7XPC, 7XRX, 7XS3, 7XSG, 7XVI, 7XWY, 7XYO, 7XZ3, 7Y01, 7Y16, 7Y1S, 7Y3Z, 7Y43, 7Y4H, 7Y7O, 7Y8H, 7Y8S, 7Y95, 7Y9A, 7Y9G, 7YDQ, 7YDW, 7YFX, 7YH1, 7YH2, 7YH5, 7YIA, 7YK3, 7YKG, 7YKM, 7YL4, 7YLQ, 7YLZ, 7YN2, 7YNX, 7YOJ, 7YPD, 7YPQ, 7YPR, 7YR9, 7YSI, 7YTU, 7YUJ, 7YUK, 7YUN, 7YYG, 7Z3I, 7Z64, 7Z8E, 7ZAO, 7ZCA, 7ZCB, 7ZDS, 7ZFR, 7ZIE, 7ZLQ, 7ZPN, 7ZRZ, 8A0A, 8A16, 8A39, 8A50, 8A82, 8A90, 8A9N, 8AA9, 8AAS, 8ABV, 8ACA, 8ACC, 8ACR, 8ADC, 8ADD, 8AEP, 8AF9, 8AGG, 8AGR, 8AHD, 8AHZ, 8AMU, 8AN0, 8ANT, 8ANU, 8ARD, 8ARE, 8AU6, 8AXF, 8AXJ, 8AXT, 8AY2, 8AYR, 8AYV, 8B2E, 8B2H, 8B2S, 8B3E, 8B3W, 8B3Z, 8B45, 8B4C, 8B4D, 8B4E, 8B5W, 8B6A, 8B6P, 8B6Q, 8B6U, 8BAT, 8BBR, 8BBZ, 8BCK, 8BDQ, 8BF8, 8BFY, 8BH8, 8BH9, 8BID, 8BJ3, 8BL6, 8BNB, 8BRD, 8BRP, 8BRY, 8BUX, 8BVE, 8BVI, 8BVK, 8BW6, 8BY5, 8BZ2, 8BZ4, 8C26, 8C27, 8C4D, 8C9V, 8CB2, 8CBV, 8CEG, 8CH4, 8CHW, 8CHX, 8CID, 8CIH, 8CIL, 8CJG, 8CK0, 8CLU, 8CLZ, 8CMQ, 8CPK, 8CPN, 8CPQ, 8CQM, 8CQO, 8CR7, 8D0P, 8D40, 8D49, 8D4A, 8D4B, 8DAJ, 8DEO, 8DHJ, 8DQ6, 8DSP, 8DTQ, 8DVQ, 8DWJ, 8DX0, 8DZJ, 8E2B, 8E55, 8E5D, 8E5E, 8E7F, 8EB5, 8EBB, 8EC3, 8EEL, 8EFZ, 8EHN, 8EHT, 8EIL, 8EIN, 8ENR, 8EOV, 8ER5, 8ERC, 8ERM, 8ESL, 8EWG, 8EX9, 8EXA, 8EZW, 8F5D, 8F5G, 8F70, 8F72, 8F7N, 8F9Y, 8FBJ, 8FBN, 8FIT, 8FJD, 8FJE, 8FKL, 8FNR, 8FTV, 8FUQ, 8FW7, 8FX9, 8FXF, 8G0N, 8G2V, 8G32, 8G49, 8G53, 8G5S, 8G8E, 8GA6, 8GAQ, 8GDW, 8GH8, 8GI1, 8GKH, 8GKV, 8GLD, 8GMX, 8GMZ, 8GOY, 8GPP, 8GPS, 8GSR, 8GSX, 8GSY, 8GTY, 8GXR, 8GY0, 8GY4, 8GYG, 8GYR, 8H0L, 8H18, 8H1J, 8H3Z, 8H55, 8H5S, 8H7A, 8H8P, 8HAW, 8HBB, 8HBR, 8HDU, 8HDV, 8HE2, 8HF2, 8HFC, 8HFP, 8HGU, 8HHL, 8HHM, 8HI8, 8HIG, 8HIO, 8HIU, 8HL8, 8HLG, 8HM4, 8HML, 8HN0, 8HNO, 8HP8, 8HR5, 8HTX, 8HUZ, 8HVC, 8I3F, 8I3J, 8I3X, 8IA3, 8IBW, 8IBX, 8IBY, 8IBZ, 8IF8, 8IFK, 8IHA, 8IIB, 8IJT, 8IKS, 8IL8, 8ILL, 8IMD, 8IMS, 8IN6, 8IN8, 8IOI, 8IOM, 8IOO, 8IP1, 8IQ8, 8IQC, 8IR2, 8IRQ, 8ISZ, 8ITN, 8ITO, 8IUB, 8IVI, 8IVZ, 8IW5, 8IWP, 8J0Q, 8J12, 8J1J, 8J3R, 8J48, 8J69, 8J8H, 8J9G, 8JAH, 8JDG, 8JFR, 8JFU, 8JHR, 8JHU, 8JI2, 8JMR, 8JNM, 8JNX, 8JP0, 8JQE, 8JU7, 8JW3, 8JWS, 8JZX, 8K1F, 8K1I, 8K1R, 8K34, 8K3D, 8K3F, 8K4L, 8K5K, 8K5U, 8K76, 8K7X, 8K7Y, 8K86, 8K8A, 8KGZ, 8OEI, 8OJN, 8OK3, 8OKH, 8OLJ, 8OML, 8OQJ, 8OU1, 8OW8, 8OWF, 8OWI, 8OXU, 8OYY, 8P0C, 8P5Q, 8P6J, 8P6M, 8PAG, 8PAQ, 8PE3, 8PHB, 8PM4, 8PMU, 8POF, 8POG, 8PR6, 8PSN, 8PSO, 8PSQ, 8PSS, 8PSU, 8PT2, 8PT6, 8PT7, 8PTH, 8PTJ, 8PX4, 8PXC, 8PZH, 8Q1K, 8Q2A, 8Q3Z, 8Q40, 8Q41, 8Q42, 8Q43, 8Q44, 8Q66, 8Q70, 8QCW, 8QDO, 8QFL, 8QK8, 8QLD, 8R79, 8R7D, 8R8A, 8RB4, 8RH3, 8RXO, 8S0U, 8S9W, 8SDZ, 8SIU, 8SM6, 8SM9, 8SME, 8SMQ, 8SOT, 8SQU, \textcolor{red}{8SRO}, 8SSC, 8SSU, 8SUT, 8SVA, 8SW0, 8SXS, 8SXU, 8SY3, 8T0P, 8T1M, 8T3J, 8T4C, 8T4R, 8T65, 8T66, 8T8L, 8TA2, 8TFS, 8TGH, 8TJG, 8TKA, 8TN8, 8TNM, 8TP8, 8TPK, 8TTO, 8TVL, 8U0M, 8U1K, 8UA5, 8UK7, 8UPI, 8UQ9, 8URN, 8UZ8, 8V14, 8V1K, 8VRM, 8VXA, 8VXC, 8W7P, 8W8S, 8W9O, 8WBR, 8WCT, 8WIL, 8WO7, 8WU2, 8X51, 8X6V, 8X7X, 8XAT, 8XBI, 8XX9, 8Y4U, 8YA7\par}

\subsection{Qualitative Results}
\label{sec:qualitatitve}

\begin{figure}[t]
  \centering
  \begin{subfigure}{\linewidth}
    \includegraphics[width=\linewidth]{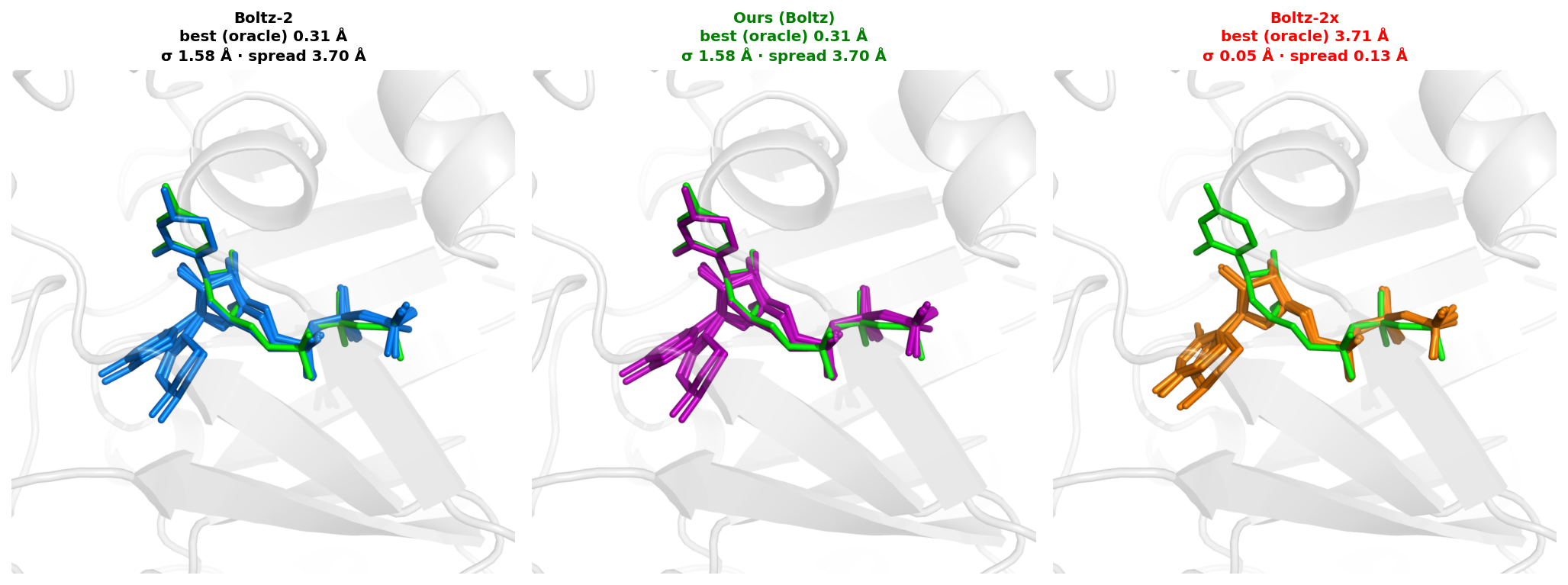}
    \caption{\textbf{7n7h}}
    \label{fig:case-7n7h}
  \end{subfigure}\\[0.6em]
  \begin{subfigure}{\linewidth}
    \includegraphics[width=\linewidth]{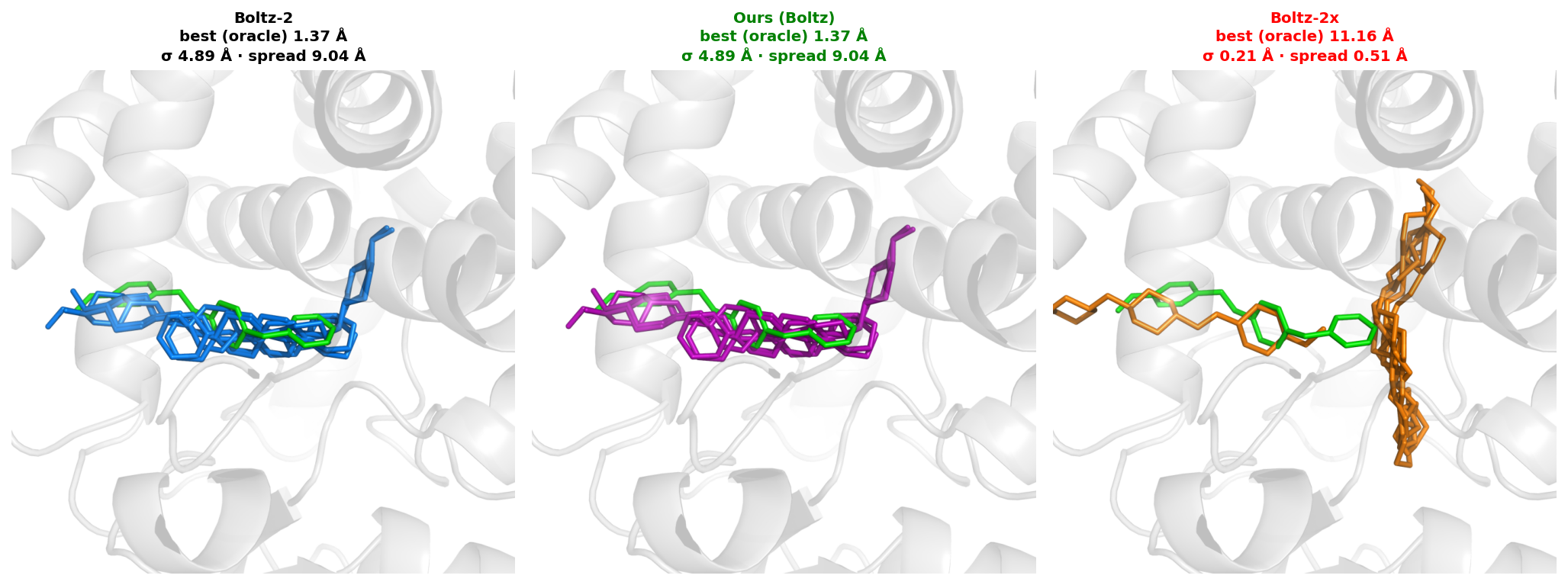}
    \caption{\textbf{7p5t}}
    \label{fig:case-7p5t}
  \end{subfigure}
  \caption{ All five sampled ligand poses per method, superposed on the crystal ligand
  (green) after alignment on the receptor. Left: Boltz-2 (blue); middle: our projection
  (magenta); right: Boltz-2x (orange). 
  Panel titles give the oracle (best-of-5) ligand RMSD and the standard deviation $\sigma$ and range (spread) of the five per-sample ligand RMSDs. Our projection reproduces the base-model ensemble almost exactly and keeps the same oracle pose, whereas steering collapses the ensemble onto a single tight but incorrect binding mode.}
  \label{fig:qualitative}
\end{figure}

Fig.~\ref{fig:qualitative} illustrates the accuracy--validity trade-off of our
approach vs.\ Boltz-2x on two PoseBusters monomer targets: 7n7h and 7p5t. On both, our projection leaves the base-model ensemble
essentially intact: the oracle RMSD is unchanged ($0.31$\,\AA\ on 7n7h, $1.37$\,\AA\ on
7p5t) while achieving physical validity. %and the sample statistics match baseline Boltz-2 to two decimal places, confirming that the
%correction acts locally on steric violations rather than displacing the ligand within the
%pocket.
Boltz-2x behaves differently. Its ensembles are often tighter
($\sigma = 0.05$ and $0.21$\,\AA\ versus $1.58$ and $4.89$\,\AA), which is the expected
consequence of importance-sampling over candidates, and
the resulting consensus pose is sometimes wrong: the oracle RMSD degrades to $3.71$\,\AA\ on 7n7h
and to $11.16$\,\AA\ on 7p5t, where every steered sample places the ligand tail outside
the crystallographic channel. This suggests that Boltz-2x steering may sometimes achieve %hard
physical validity at the expense of prediction quality, %structural quality,
reweighting the sampler decisively towards a physically clean but biologically incorrect
binding mode. %This is the qualitative counterpart of the Boltz-2x ligand-placement deficits in
%Table~\ref{tab:main_results}.

\section{Time and Memory}

\begin{table}[t]
\centering
\caption{Per-target inference cost on five datasets. Runtime is in seconds and peak GPU memory in GiB, both over five samples: median, interquartile range (IQR, p25--p75) and mean or maximum over targets. Rel.\ is the median over targets of the per-target runtime ratio to the method's base model (Boltz-2 for Boltz-2x and Ours (Boltz); OpenFold-3 for Ours (OpenFold)). Runtimes are comparable within a model family, not across families. Only targets completed by every method are included.}
\label{tab:time_mem_combined}
\scriptsize
\setlength{\tabcolsep}{3.5pt}
\begin{tabular}{@{}l r r c r r r c r@{}}
\toprule
 & & \multicolumn{4}{c}{Runtime (s)} & \multicolumn{3}{c}{Peak GPU memory (GiB)} \\
\cmidrule(lr){3-6} \cmidrule(l){7-9}
Method & $n$ & Median & IQR & Mean & Rel. & Median & IQR & Max \\
\midrule
\multicolumn{9}{c}{\textbf{CASP15}} \\
\midrule
Boltz-2 & 66 & 60.3 & 22.8--148 & 193 & -- & 5.7 & 2.8--10.2 & 31.8 \\
\textbf{Ours (Boltz)} & 66 & 58.0 & 23.6--146 & 193 & 1.00 & 5.7 & 2.8--10.2 & 31.8 \\
Boltz-2x & 66 & 120 & 65.1--266 & 302 & 1.77 & 6.0 & 3.4--10.7 & 59.1 \\
OpenFold-3 & 66 & 28.3 & 24.0--43.9 & 55.5 & -- & 7.7 & 2.8--16.6 & 25.1 \\
\textbf{Ours (OpenFold)} & 66 & 28.2 & 23.9--47.0 & 56.0 & 1.01 & 7.7 & 2.8--16.6 & 25.1 \\
\midrule
\multicolumn{9}{c}{\textbf{CASP16}} \\
\midrule
Boltz-2 & 52 & 65.4 & 26.4--167 & 201 & -- & 5.3 & 2.6--10.6 & 33.6 \\
\textbf{Ours (Boltz)} & 52 & 67.4 & 27.8--165 & 201 & 1.00 & 5.3 & 2.6--10.6 & 33.6 \\
Boltz-2x & 52 & 120 & 54.4--285 & 318 & 1.78 & 6.4 & 2.9--11.4 & 53.0 \\
OpenFold-3 & 52 & 32.1 & 24.8--49.2 & 58.1 & -- & 9.5 & 3.5--12.5 & 30.7 \\
\textbf{Ours (OpenFold)} & 52 & 31.4 & 24.1--51.4 & 57.6 & 0.99 & 9.5 & 3.5--12.5 & 30.7 \\
\midrule
\multicolumn{9}{c}{\textbf{PoseBusters Monomer}} \\
\midrule
Boltz-2 & 134 & 49.4 & 37.9--70.9 & 67.5 & -- & 7.4 & 5.5--9.4 & 14.8 \\
\textbf{Ours (Boltz)} & 134 & 50.4 & 39.1--72.6 & 68.2 & 1.01 & 7.4 & 5.5--9.4 & 14.8 \\
Boltz-2x & 134 & 106 & 79.9--147 & 132 & 2.05 & 7.4 & 5.5--9.4 & 14.8 \\
OpenFold-3 & 134 & 26.1 & 24.4--30.2 & 30.2 & -- & 6.4 & 4.5--9.4 & 23.8 \\
\textbf{Ours (OpenFold)} & 134 & 27.1 & 24.8--31.0 & 30.8 & 1.02 & 6.4 & 4.5--9.4 & 23.8 \\
\midrule
\multicolumn{9}{c}{\textbf{PoseBusters Complex}} \\
\midrule
Boltz-2 & 164 & 170 & 91.5--345 & 295 & -- & 14.8 & 9.3--22.1 & 44.3 \\
\textbf{Ours (Boltz)} & 164 & 171 & 91.6--349 & 296 & 1.01 & 14.8 & 9.3--22.1 & 44.3 \\
Boltz-2x & 164 & 294 & 177--574 & 493 & 1.75 & 16.5 & 9.7--23.8 & 78.2 \\
OpenFold-3 & 164 & 49.1 & 34.6--85.9 & 72.4 & -- & 13.2 & 9.9--18.2 & 33.1 \\
\textbf{Ours (OpenFold)} & 164 & 47.5 & 33.4--84.8 & 71.0 & 1.00 & 13.2 & 9.9--18.2 & 33.1 \\
\midrule
\multicolumn{9}{c}{\textbf{Test}} \\
\midrule
Boltz-2 & 541 & 65.1 & 41.7--191 & 178 & -- & 6.4 & 3.9--9.8 & 36.8 \\
\textbf{Ours (Boltz)} & 541 & 67.3 & 41.6--193 & 177 & 1.00 & 6.4 & 3.9--9.8 & 36.8 \\
Boltz-2x & 541 & 144 & 72.6--341 & 296 & 1.77 & 7.2 & 4.2--12.2 & 75.3 \\
OpenFold-3 & 541 & 29.8 & 23.5--52.9 & 48.4 & -- & 9.2 & 4.1--14.7 & 24.3 \\
\textbf{Ours (OpenFold)} & 541 & 30.8 & 24.9--54.5 & 49.8 & 1.03 & 9.2 & 4.1--14.7 & 24.3 \\
\bottomrule
\end{tabular}
\end{table}

Table~\ref{tab:time_mem_combined} reports the inference cost of every method on NVIDIA A100-SXM4-80GB. The projection adds almost nothing: \textbf{Ours (Boltz)} runs at 1.001--1.007$\times$ its time per target (median ratio per dataset), and its peak GPU memory is unchanged (largest median per-target difference 0.00~GiB), because it operates on coordinates the sampler already holds and adds no network evaluations. \textbf{Ours (OpenFold)} behaves the same way relative to OpenFold-3 (0.99--1.03$\times$ its time, memory unchanged). Inference-time steering is far more expensive: Boltz-2x takes roughly 1.7--2.1$\times$ the time of Boltz-2, needs up to 78~GiB where Boltz-2 needs at most 44~GiB on the same targets, and runs out of memory on 9 targets that Boltz-2 and Ours (Boltz) both complete. The absolute runtimes of the two model families are not directly comparable, since the OpenFold timer covers only the predict step.

\section{Structural effects of the projection}
\label{app:structural_effects}

The projection moves atoms, so a natural concern is that it resolves steric
violations at the expense of other aspects of structural quality. We test this
in three ways: we compare in-loop projection with post-hoc correction
(\S\ref{app:posthoc}), and we check covalent geometry
(\S\ref{app:covalent}), 
% backbone and side-chain conformations
% (\S\ref{app:conformations}),
and atomic packing and non-bonded energies
(\S\ref{app:packing}). All three point to the same conclusion: the projection
removes steric violations while leaving accuracy, covalent geometry, and local packing statistically indistinguishable
from the base model.

%------------------------------------------------------------------
\subsection{In-loop projection versus post-hoc correction}
\label{app:posthoc}

We apply the projection inside the sampler rather than to the final
structure. To test whether this matters, we compare it with two post-hoc
alternatives applied to the output of the unmodified model.

\paragraph{Post-hoc projection.}
We apply the same operator $\Pi_{\mathcal{C}(\mathcal{B}_{\mathrm{proj}})}$,
with identical hyperparameters, to final Boltz-2 structures, either once or
iterated until no targeted violation remains (40 passes). A single
pass removes only $14\%$ of the targeted violations and raises physical
validity to $0.22$, compared with $1.00$ for in-loop projection
(Table~\ref{tab:sen_position}). This is expected, because each pass moves an
atom by at most $\delta_{\max}$. Iterating to feasibility recovers validity 
($0.90$) but significantly lowers LDDT ($0.79$ versus $0.81$, Holm-corrected
paired test). Once sampling is over, the denoiser can no longer absorb the
corrections, so they are imposed on the structure instead of being integrated
into it, hence, affecting the structural quality.

\paragraph{Post-hoc geometry minimization.}
We also relax each predicted structure with
\texttt{phenix.geometry\_minimization} \citep{phenix}, which minimizes against
a standard stereochemical restraint library covering bonds, angles,
dihedrals, planarity, chirality and non-bonded clash terms.  
Phenix could not
generate restraints for ligand-bearing targets, so this comparison covers the
42 of 52 CASP16 targets without ligands. Minimization improves validity from
$0.595$ to $0.881$ but lowers accuracy on every per-atom metric (except DockQ)
(Table~\ref{tab:comparison_minim}). Our projection instead reaches $1.000$
validity ($+0.405$ [$+0.262$, $+0.548$], $p = 1.2\times10^{-4}$) while
remaining statistically indistinguishable from the base model on every
structural metric (all n.s., $|\Delta| \le 0.003$ for LDDT, backbone LDDT and
TM-score). Head-to-head, projection significantly outperforms minimization on
LDDT ($+0.022$, $p = 7.4\times10^{-5}$) and backbone LDDT ($+0.027$,
$p = 1.6\times10^{-6}$). The accuracy loss from minimization is also largest
where the prediction is best. Per-target LDDT change correlates at $-0.57$
with baseline accuracy: it averages $+0.001$ for models below $0.4$ LDDT,
$-0.020$ for $0.4$--$0.7$, and $-0.052$ above $0.7$. Post-hoc minimization
therefore trades accuracy for validity, and it does so most on the models
that are already accurate. In-loop projection avoids this trade-off and can be applied on all types of biomolecules.

\textbf{Protocol:} Each structure is minimized for three macro-cycles with the built-in monomer
library. We record the RMSD between the input and minimized structure and verified that minimization did not move atoms by more than a size-dependent RMSD threshold. The threshold ($\mathrm{RMSD_{thresh}}$)  scales with the size of the query protein \citep{Carugo2001-yj}, defined as:
    
    \begin{align}
    \label{eq:RMSD_threhsold}
    \mathrm{RMSD_{thresh}} = \left(1 + \ln\sqrt{\frac{N}{100}}\right)\mathrm{RMSD}_{100}
    \end{align}
    
where $N$ is the size of the protein and RMSD\textsubscript{100} is a selected reference threshold for a protein of length 100 residues, selected here to be 1.0{\AA}. We found that none of the structures exceeded the corresponding threshold. All comparisons use the oracle model (best of five) and paired Wilcoxon
signed-rank tests with Holm correction.

\begin{table}[t]
\centering
\caption{In-loop projection versus post-hoc Phenix minimization on the 42
CASP16 targets for which Phenix can build restraints (ligand-free). Oracle
(best-of-five) model per target. Arrows indicate the direction of
improvement; DockQ$_{>0.23}$ is computed over the 17 multi-chain targets.}
\label{tab:comparison_minim}
\footnotesize
\setlength{\tabcolsep}{6pt}
\begin{tabular}{@{}lrrr@{}}
\toprule
Metric & Boltz-2 & Phenix min.\ & Ours (Boltz) \\
\midrule\midrule
Physically valid $\uparrow$         & 0.595          & 0.881 & \textbf{1.000} \\
LDDT $\uparrow$                     & \textbf{0.385} & 0.363 & \textbf{0.385} \\
Backbone LDDT $\uparrow$            & \textbf{0.422} & 0.395 & \textbf{0.422} \\
TM-score $\uparrow$                 & 0.394          & 0.391 & \textbf{0.396} \\
Global RMSD (\AA) $\downarrow$      & 14.7           & 14.8  & \textbf{14.6}  \\
DockQ$_{>0.23}$ $\uparrow$ ($n=17$) & 0.510          & 0.510 & \textbf{0.569} \\
\bottomrule
\end{tabular}
\end{table}

\subsection{Covalent geometry}
\label{app:covalent}

\begin{figure}[t]
    \centering
    \includegraphics[width=\textwidth]{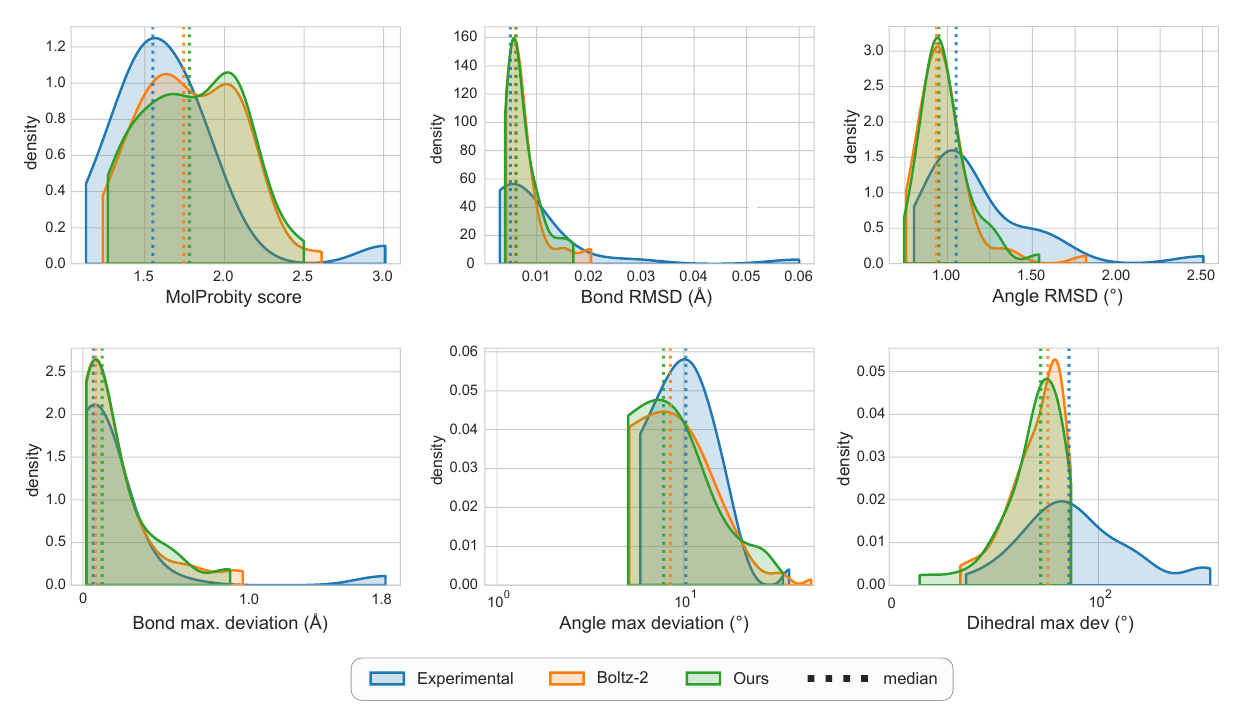}
    \caption{\textbf{MolProbity covalent-geometry metrics on CASP15 and CASP16
    protein chains.} Kernel density estimates across targets for the
    experimental structures (blue), Boltz-2 (orange) and our projection
    (green): MolProbity score, bond and angle RMSD, and maximum bond, angle
    and dihedral deviation. Dotted lines mark the medians. The angle maximum
    deviation is shown on a logarithmic axis.}
    \label{fig:bounded_geometries}
\end{figure}

The van der Waals projection moves atoms without reference to their bonds, so
it could in principle stretch bonds or distort angles. We assess covalent
geometry of the protein chains with MolProbity  \cite{Davis2007-nq}
(Fig.~\ref{fig:bounded_geometries}, Table~\ref{tab:covalent_geometry}).
Bonded geometry is essentially unchanged by the projection. The median bond
RMSD from ideal is $0.0062$\,\AA{} with projection and $0.0060$\,\AA{} for
Boltz-2, against $0.0050$\,\AA{} for the experimental structures, i.e.\
${\sim}0.4\%$ of a $1.52$\,\AA{} C--C bond. The median angle RMSD is
$0.955^\circ$ versus $0.940^\circ$ (experimental $1.051^\circ$), or
${\sim}0.85\%$ of a tetrahedral angle. The worst-case angle and dihedral
deviations are slightly lower with projection (median $7.32^\circ$ vs.\
$8.13^\circ$ and $78.4^\circ$ vs.\ $80.8^\circ$). The one metric that moves
upward is the worst-case bond deviation (median $0.114$ vs.\ $0.070$\,\AA{}); its upper tail is lower with projection
(p$_{95}$ $0.568$ vs.\ $0.625$\,\AA{}). 
Both predicted models are closer to ideal than the deposited structures on angle RMSD ($0.955^\circ$ and $0.940^\circ$  versus $1.051^\circ$), while their bond RMSD is marginally higher (0.0062\AA{} and 0.0060\AA{} versus 0.0050\AA{}). The angle result likely reflects the bias of cofolding models toward the idealized geometry learned during training. The overall MolProbity score is essentially unchanged (median 1.780 with projection versus 1.744 for Boltz-2, against 1.550 for the experimental structures).

\paragraph{Protocol:}
We run \texttt{phenix.molprobity} (\cite{phenix}, Phenix 2.1-6048) on the CASP15 and CASP16
protein chains. Structures are first prepared with \texttt{phenix.pdbtools},
removing waters, ligands and ions and keeping only the primary alternate
conformation (selection \texttt{not water and (altloc " " or altloc A) and
protein}). For large complexes, where \texttt{pdbtools} writes mmCIF, the
mmCIF file is passed to MolProbity directly. Hydrogens are added with
\texttt{reduce}, which optimizes the hydrogen-bond network and flips
Asn/Gln/His side chains where needed, so that all-atom contacts are computed
on fully protonated models. Bond and angle deviations are evaluated against
the standard monomer (geostd) library. Each target contributes the mean over
its five predicted models; the experimental structure contributes once.
Because ligands are removed, ligand-internal geometry is assessed only by the
validity criterion.

\begin{table}[t]
\centering
\caption{Covalent geometry and overall quality of CASP protein chains
($n=46$). Each target contributes its mean over the five predicted models;
the table reports the distribution across targets as median [Q$_1$, Q$_3$].
Lower is better throughout ($\downarrow$).}
\label{tab:covalent_geometry}
\footnotesize
\setlength{\tabcolsep}{5pt}
\begin{tabular}{@{}lccc@{}}
\toprule
Metric (unit) $\downarrow$ & Experimental & Boltz-2 & Ours \\
\midrule\midrule
MolProbity score                & 1.550 [1.480, 1.780]     & 1.744 [1.552, 2.036]     & 1.780 [1.528, 2.064]     \\
Bond RMSD (\AA)                 & 0.0050 [0.0050, 0.0080]  & 0.0060 [0.0052, 0.0076]  & 0.0062 [0.0052, 0.0084]  \\
Angle RMSD ($^\circ$)           & 1.051 [1.027, 1.278]     & 0.940 [0.896, 1.003]     & 0.955 [0.908, 1.031]     \\
Bond max.\ dev.\ (\AA)          & 0.061 [0.048, 0.085]     & 0.070 [0.036, 0.207]     & 0.114 [0.034, 0.240]     \\
Angle max.\ dev.\ ($^\circ$)    & 10.310 [8.747, 17.225]   & 8.128 [5.349, 16.763]    & 7.323 [5.406, 18.196]    \\
Dihedral max.\ dev.\ ($^\circ$) & 88.36 [84.64, 110.78]    & 80.79 [73.85, 83.88]     & 78.42 [73.94, 82.80]     \\
\bottomrule
\end{tabular}
\end{table}

\subsection{Packing and non-bonded energetics}
\label{app:packing}

Finally, we check that the projection does not disturb the local environment
of each residue. We use two independent measures on the CASP16 systems:
Lennard-Jones energies under a molecular-mechanics force field, and contact
counts in the alpha complex.

\paragraph{Lennard-Jones energy.}
Before relaxation, the per-residue Lennard-Jones energies of the two methods
are similar on most targets, with a lower median under projection ($224.5$
vs.\ $249.5$\,kJ/mol per residue; Table~\ref{tab:pot_energy}). The
distribution is heavy-tailed. On a few targets a single severe overlap
dominates the Lennard-Jones term, and these are exactly the targets where the
projection helps most: on T1272s5, for example, the energy drops from
${\sim}6.4\times10^{6}$ to ${\sim}1.1\times10^{4}$\,kJ/mol per residue. This is
why the means differ by more than two orders of magnitude. After an
identical short relaxation, the two methods reach the same energy (median
$145.1$ vs.\ $145.5$\,kJ/mol per residue), agreeing to within a fraction of a
kJ/mol on every target. The projection therefore removes the severe contacts
without moving the structure away from the same local energy minimum that the
base model reaches.

\paragraph{Alpha-complex packing.}
The number of non-covalent neighbours per residue in the solvent-accessible
alpha complex is essentially unchanged (Table~\ref{tab:alpha_complex}): the
median per-target change is $+0.003$ neighbours per residue
($\le 0.01\%$), and per-target differences are distributed symmetrically
around zero. The largest changes occur on RNA and large multimeric targets,
and even these stay below $2\%$ and come with no degradation in covalent
geometry or energy. The projection removes clashes without changing the
contact topology of the structure.

\paragraph{Protocol.}
Lennard-Jones energies are computed with OpenMM \citep{openmm} and the AMBER
ff14SB force field. We evaluate each raw prediction
(pre-relaxation) and the same structure after an identical minimization of
100 steps with ff14SB and GBn2 implicit solvent (post-relaxation). The
relaxation removes the artefactual energy of a few severe initial contacts and
evaluates the structure in a nearby low-energy state. The alpha complex is
built with GUDHI \citep{gudhi}, which uses the CGAL alpha-shape
implementation, on atoms with van der Waals radii inflated by a $1.4$\,\AA{}
solvent probe, at $\alpha = 0$. Two atoms are neighbours when their inflated
spheres overlap without occlusion; covalently bonded pairs are excluded.

\begin{table}[t]
  \centering
  \caption{Packing on CASP16 systems for Boltz-2 and our projection.
  $\downarrow$: lower is better.}
  \label{tab:vdw}
  \begin{subtable}{\linewidth}
    \centering
    \caption{Lennard-Jones energy per residue (kJ/mol) before and after an
    identical short relaxation ($\downarrow$).}
    \label{tab:pot_energy}
    \begin{tabular}{lrrrr}
      \toprule
             & \multicolumn{2}{c}{Boltz-2} & \multicolumn{2}{c}{Ours} \\
      \cmidrule(lr){2-3} \cmidrule(lr){4-5}
      LJ energy $\downarrow$ & Pre & Post & Pre & Post \\
      \midrule
      Mean   & 212440.8 & 146.0 & 1216.8 & 146.0 \\
      Median & 249.5    & 145.1 & 224.5  & 145.5 \\
      \bottomrule
    \end{tabular}
  \end{subtable}

  \vspace{0.8em}

  \begin{subtable}{\linewidth}
    \centering
    \caption{Non-covalent neighbour counts in the solvent-accessible alpha
    complex ($\alpha=0$), reported as median [Q$_1$, Q$_3$] (max). Neither
    direction is better: the test is whether the projection leaves the counts
    unchanged relative to the base model.}
    \label{tab:alpha_complex}
    \begin{tabular}{lcc}
      \toprule
      Metric (no. of neighbours) & Boltz-2 & Ours \\
      \midrule
      Neighbours per residue & 45.84 [42.97, 99.85] (111.35) & 45.90 [42.99, 99.84] (110.85) \\
      Neighbours (total)     & 20700 [10216, 38518] (68036)  & 20744 [10217, 38530] (68175) \\
      \bottomrule
    \end{tabular}
  \end{subtable}
\end{table}

\section{Hyperparameter Sensitivity}

The sweeps in Tables ~\ref{tab:sen_position} and ~\ref{tab:sensitivity} use a pooled
set of $134$ targets drawn from CASP15, CASP16, PoseBusters Monomer and
PoseBusters Complex: every target on which the unmodified Boltz-2 sampler
produced at least one physically invalid sample. 
% We restrict the sweep to
% these targets because they are the only ones on which the projection can
% % make a difference; on targets the base model already gets right, every
% setting scores identically and the comparison carries no information. 
One setting is
varied at a time; all others stay at the final configuration.
$\delta_{\max}$ (\texttt{projection\_max\_atom\_disp}, \AA) caps the net
per-atom displacement; $\alpha$ (\texttt{projection\_alpha}) scales the
per-pair correction; $\beta$ (\texttt{vdw\_projection\_buffer}) sets the vdW
lower bound $(1-\beta)(r_i+r_j)$; $K$ (\texttt{projection\_max\_pairs}) caps
the violating pairs corrected per step; $t_{\mathrm{start}}$
(\texttt{projection\_start\_t}) opens a projection window that runs to the end
of sampling.

% Two
% consequences follow for reading the tables. First, physical validity under the
% base model is $0$ on this set by construction, so the validity column measures
% how completely a setting repairs known failures rather than performance on a
% representative benchmark; . Second, 

Because the set is enriched for hard,
clash-heavy structures, the accuracy columns are lower than the
benchmark averages and are best read as relative comparisons between rows,
not as absolute accuracy. Full benchmark-level comparison is available in main text Table~\ref{tab:main_results}.
% Accuracy metrics are computed on the subset of
% targets for which each is defined.

\begin{table}[t]
\centering
\footnotesize
\caption{Sensitivity of the projection to its hyperparameters on the
sensitivity set (CASP15, CASP16, PoseBusters Monomer and Complex). Values are
mean $\pm$ non-parametric bootstrap standard error over targets (10{,}000
resamples), oracle over 5 samples, computed as in
Table~\ref{tab:main_results}. Higher is better for all metrics.  Dashed, shaded rows are the final configuration used in the main
experiments. Red cells are worse than the final configuration under a paired
test (Wilcoxon; exact McNemar for physical validity) surviving Holm
correction over the table.}
\label{tab:sensitivity}
\setlength{\tabcolsep}{4pt}
\begin{tabular}{@{}l c c c c c c@{}}
\toprule
Parameter & Value & LDDT & DockQ$_{>0.23}$ & LDDT-PLI & LRMSD$_{<2}$ & Phys.\ valid \\
\midrule
$\delta_{\max}$ (\AA) & 0.01 & $0.81 \pm 0.02$ & $0.55 \pm 0.04$ & $0.54 \pm 0.06$ & $0.53 \pm 0.08$ & \cellcolor{red!15}$0.27 \pm 0.04$ \\
$\delta_{\max}$ (\AA) & 0.15 & $0.81 \pm 0.02$ & $0.55 \pm 0.04$ & $0.53 \pm 0.06$ & $0.52 \pm 0.08$  & $1.00 \pm 0.00$\\
$\delta_{\max}$ (\AA) & 0.30 & $0.81 \pm 0.02$ & $0.55 \pm 0.04$ & $0.53 \pm 0.06$ & $0.52 \pm 0.08$  & $1.00 \pm 0.00$ \\
\hdashline
\rowcolor{green!12}$\delta_{\max}$ (\AA) & 0.45 & $0.81 \pm 0.02$ & $0.55 \pm 0.04$ & $0.54 \pm 0.06$ & $0.53 \pm 0.08$  & $1.00 \pm 0.00$ \\
\hdashline
$\delta_{\max}$ (\AA) & 0.60 & $0.81 \pm 0.02$ & $0.55 \pm 0.04$ & $0.54 \pm 0.06$ & $0.53 \pm 0.08$  & $1.00 \pm 0.00$\\
$\delta_{\max}$ (\AA) & 0.90 & $0.81 \pm 0.02$ & $0.55 \pm 0.04$ & $0.54 \pm 0.06$ & $0.53 \pm 0.08$  & $1.00 \pm 0.00$ \\
\midrule\midrule
$\alpha$ & 0.10 & $0.81 \pm 0.02$ & $0.55 \pm 0.04$ & $0.53 \pm 0.06$ & $0.52 \pm 0.08$ & \cellcolor{red!15}$0.51 \pm 0.04$ \\
\hdashline
\rowcolor{green!12}$\alpha$ & 0.25 & $0.81 \pm 0.02$ & $0.55 \pm 0.04$ & $0.54 \pm 0.06$ & $0.53 \pm 0.08$  & $1.00 \pm 0.00$ \\
\hdashline
$\alpha$ & 0.50 & $0.81 \pm 0.02$ & $0.55 \pm 0.04$ & $0.54 \pm 0.06$ & $0.53 \pm 0.08$  & $1.00 \pm 0.00$\\
$\alpha$ & 0.75 & $0.81 \pm 0.02$ & $0.55 \pm 0.04$ & $0.53 \pm 0.06$ & $0.52 \pm 0.08$  & $1.00 \pm 0.00$ \\
$\alpha$ & 1.00 & $0.81 \pm 0.02$ & $0.55 \pm 0.04$ & $0.53 \pm 0.06$ & $0.52 \pm 0.08$  & $1.00 \pm 0.00$ \\
\midrule\midrule
$\beta$ & 0.01 & \cellcolor{red!15}$0.80 \pm 0.02$ & $0.55 \pm 0.04$ & $0.52 \pm 0.06$ & $0.49 \pm 0.08$ & $0.93 \pm 0.02$ \\
$\beta$ & 0.10 & \cellcolor{red!15}$0.80 \pm 0.02$ & $0.55 \pm 0.04$ & $0.54 \pm 0.06$ & $0.53 \pm 0.08$  & $1.00 \pm 0.00$ \\
\hdashline
\rowcolor{green!12}$\beta$ & 0.15 & $0.81 \pm 0.02$ & $0.55 \pm 0.04$ & $0.54 \pm 0.06$ & $0.53 \pm 0.08$  & $1.00 \pm 0.00$ \\
\hdashline
$\beta$ & 0.50 & $0.81 \pm 0.02$ & $0.55 \pm 0.04$ & $0.54 \pm 0.06$ & $0.53 \pm 0.08$ & \cellcolor{red!15}$0.01 \pm 0.01$ \\
$\beta$ & 0.95 & $0.81 \pm 0.02$ & $0.55 \pm 0.04$ & $0.54 \pm 0.06$ & $0.53 \pm 0.07$ & \cellcolor{red!15}$0.01 \pm 0.01$ \\
\midrule\midrule
$K$ & 1 & $0.81 \pm 0.02$ & $0.55 \pm 0.04$ & $0.54 \pm 0.06$ & $0.53 \pm 0.08$ & \cellcolor{red!15}$0.51 \pm 0.04$ \\
$K$ & 10 & $0.81 \pm 0.02$ & $0.55 \pm 0.04$ & $0.54 \pm 0.06$ & $0.53 \pm 0.08$ & $0.96 \pm 0.02$ \\
$K$ & 50 & $0.81 \pm 0.02$ & $0.55 \pm 0.04$ & $0.54 \pm 0.06$ & $0.53 \pm 0.08$  & $1.00 \pm 0.00$ \\
$K$ & 100 & $0.81 \pm 0.02$ & $0.55 \pm 0.04$ & $0.54 \pm 0.06$ & $0.53 \pm 0.08$  & $1.00 \pm 0.00$ \\
\hdashline
\rowcolor{green!12}$K$ & 200 & $0.81 \pm 0.02$ & $0.55 \pm 0.04$ & $0.54 \pm 0.06$ & $0.53 \pm 0.08$ &   $1.00 \pm 0.00$ \\
\midrule\midrule
$t_{\mathrm{start}}$ & 0.01 & $0.81 \pm 0.02$ & $0.55 \pm 0.04$ & $0.53 \pm 0.06$ & $0.52 \pm 0.08$ & \cellcolor{red!15}$0.43 \pm 0.04$ \\
$t_{\mathrm{start}}$ & 0.05 & \cellcolor{red!15}$0.80 \pm 0.02$ & $0.55 \pm 0.04$ & $0.54 \pm 0.06$ & $0.53 \pm 0.08$ & \cellcolor{red!15}$0.90 \pm 0.03$ \\
\hdashline
\rowcolor{green!12}$t_{\mathrm{start}}$ & 0.20 & $0.81 \pm 0.02$ & $0.55 \pm 0.04$ & $0.54 \pm 0.06$ & $0.53 \pm 0.08$  & $1.00 \pm 0.00$ \\
\hdashline
$t_{\mathrm{start}}$ & 0.40 & $0.81 \pm 0.02$ & $0.55 \pm 0.04$ & $0.53 \pm 0.06$ & $0.52 \pm 0.08$ & $1.00 \pm 0.00$ \\
$t_{\mathrm{start}}$ & 0.80 & $0.81 \pm 0.02$ & $0.55 \pm 0.04$ & $0.53 \pm 0.06$ & $0.52 \pm 0.08$ & $1.00 \pm 0.00$ \\
$t_{\mathrm{start}}$ & 1.00 & $0.81 \pm 0.02$ & $0.55 \pm 0.04$ & $0.53 \pm 0.06$ & $0.52 \pm 0.08$ & $1.00 \pm 0.00$ \\
\bottomrule
\end{tabular}
\end{table}

\end{document}